\documentclass[letterpaper, 10 pt, journal, twoside]{IEEEtran}

\usepackage{times}
\usepackage{epsfig}
\usepackage{graphicx}
\usepackage{amsmath}
\usepackage{amssymb}
\usepackage{amsfonts}
\usepackage{multirow}
\usepackage{booktabs}    
\usepackage{chngpage}
\usepackage{xcolor}
\usepackage{soul}
\usepackage{hyperref}
\usepackage[flushleft,para]{threeparttable}

\hypersetup{
    colorlinks=true,
    linkcolor=blue,    
    urlcolor=cyan
}

\newcommand{\real}{\mathbb{R}}

\newcommand{\change}[1]{ {\color{black}#1} }

\newcommand{\figref}[1]{Fig.~\ref{#1}}
\newcommand{\tabref}[1]{Table~\ref{#1}}
\newcommand{\secref}[1]{Sec.~\ref{#1}}
\newcommand{\eqnref}[1]{Eqn.~\ref{#1}}

\newcommand{\void}{\text{VOID }}

\begin{document}

\title{Unsupervised Depth Completion from \\ Visual Inertial Odometry}

\author{Alex Wong$^\dagger$, Xiaohan Fei$^\dagger$, Stephanie Tsuei, and Stefano Soatto
\thanks{Manuscript received: September 10, 2019; Revised December 4, 2019; Accepted January 8, 2020.}
\thanks{This paper was recommended for publication by Editor Cesar Cadena upon evaluation of the Associate Editor and Reviewers' comments.
This work was supported by ONR N00014-19-1-2066, and ONR N00014-19-1-2229.} 
\thanks{The authors are with the Samueli School of Engineering, Computer Science Department, University of California, Los Angeles, USA. Email: \texttt{\small \{alexw, feixh, 
stephanietsuei, soatto\}@cs.ucla.edu}}%
\thanks{$^\dagger$ denotes authors with equal contributions.}%
\thanks{Digital Object Identifier (DOI): see top of this page.}
}

\markboth{IEEE Robotics and Automation Letters. Preprint Version. Accepted January, 2020}
{Wong \MakeLowercase{\textit{et al.}}: Unsupervised Depth Completion from Visual Inertial Odometry} 

\maketitle

\begin{abstract}
We describe a method to infer dense depth from camera motion and sparse depth as estimated using a visual-inertial odometry system. Unlike other scenarios using point clouds from lidar or structured light sensors, we have few hundreds to few thousand points, insufficient to inform the topology of the scene. Our method first constructs a piecewise planar scaffolding of the scene, and then uses it to infer dense depth using the image along with the sparse points. We use a predictive cross-modal criterion, akin to ``self-supervision,'' measuring photometric consistency across time, forward-backward pose consistency, and geometric compatibility with the sparse point cloud.  We also present the first visual-inertial + depth dataset, which we hope will foster additional exploration into combining the complementary strengths of visual and inertial sensors. To compare our method to prior work, we adopt the unsupervised KITTI depth completion benchmark, where we achieve state-of-the-art performance. Code available at:  \hyperlink{https://github.com/alexklwong/unsupervised-depth-completion-visual-inertial-odometry}{https://github.com/alexklwong/unsupervised-depth-completion-visual-inertial-odometry}.
\end{abstract}


\IEEEpeerreviewmaketitle

\section{Introduction}
%
%
%
%
\IEEEPARstart{A}{} sequence of images is a rich source of information about both the three-dimensional (3D) shape of the environment and the motion of the sensor within. Motion can be inferred at most up to a scale and a global Euclidean reference frame, provided sufficient parallax and a number of visually discriminative Lambertian regions that are fixed in the environment and visible from the camera. The position of such regions in the scene defines the Euclidean reference frame, with respect to which motion is estimated. Scale, as well as two directions of orientation, can be further identified by fusion with inertial measurements (accelerometers and gyroscopes) and, if available, a magnetometer can fix the last (Gauge) degree of freedom.

\begin{figure}[ht]
    \centering
    \includegraphics[width=0.50\textwidth]{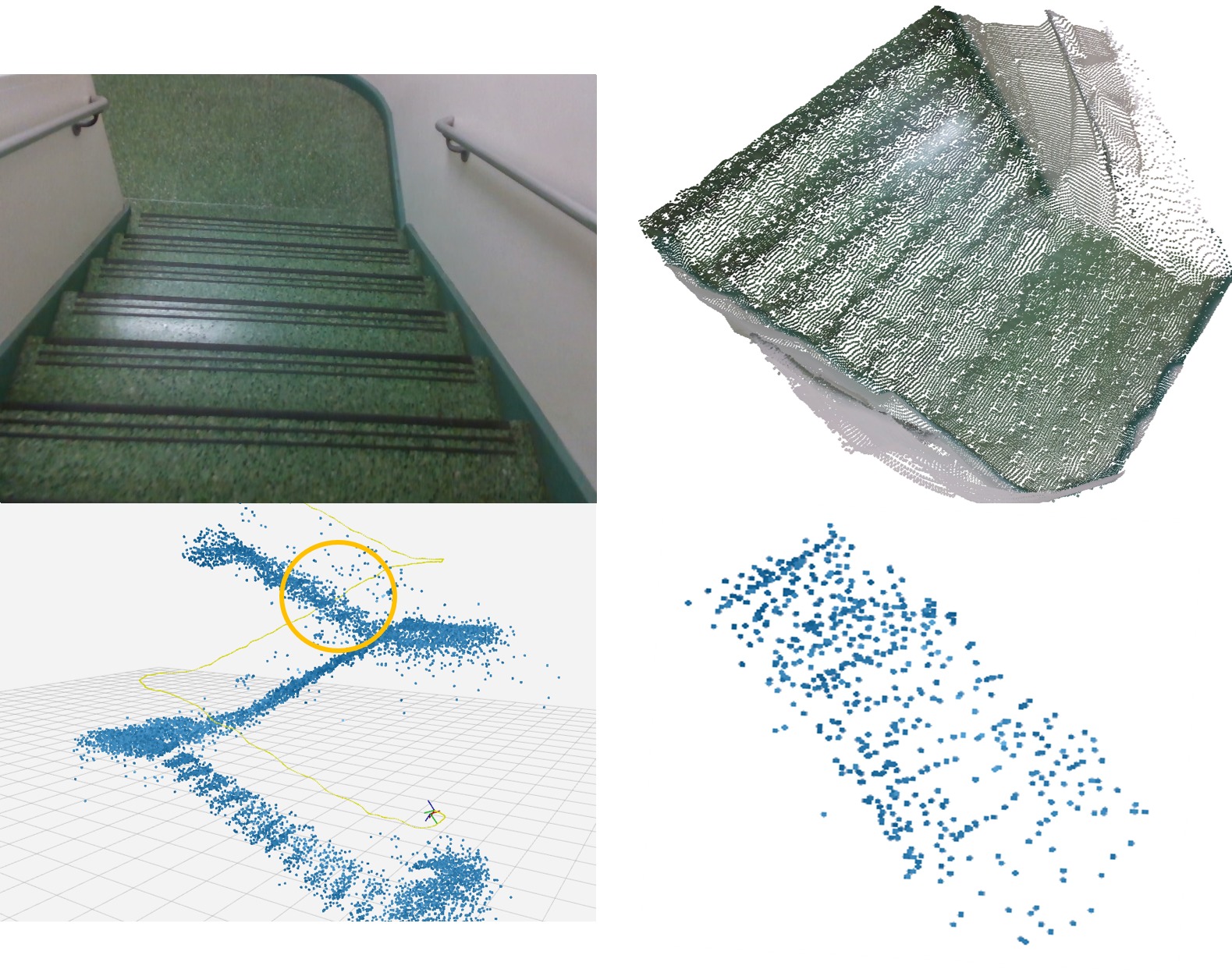}
    \vspace{-2.5em}
    \caption{\textit{Depth completion with Visual-Inertial Odometry (VIO) on the proposed VOID dataset} (best viewed in color at $5\times$). Bottom left: sparse reconstruction (blue) and camera trajectory (yellow) from VIO. The highlighted region is densified and zoomed in on the top right. Top left shows an image of the same region which is taken as input, and fused with the sparse depth image by our method. On the bottom right is the same view showing only the sparse points, insufficient to determine scene geometry and topology.}
    \label{fig:teaser}
    \vspace{-2em}
\end{figure}

Because the regions defining the reference frame have to be visually distinctive (``features''), they are typically {\em sparse}. In theory, three points are sufficient to define a Euclidean Gauge if visible at all times. In practice, because of occlusions, any Structure From Motion (SFM) or simultaneous localization and mapping (SLAM) system maintains an estimate of the location of a sparse set of features, or ``sparse point cloud,'' typically in the hundreds to thousands. These are sufficient to support a point-estimate of motion, but a rather poor representation of shape as they do not reveal the topology of the scene: The empty space between points could be empty, or occupied by a solid with a smooth surface radiance (appearance). Attempts to {\em densify} the sparse point cloud, by interpolation or regularization with generic priors such as smoothness, piecewise planarity and the like, typically fail since SFM yields far too sparse a reconstruction to inform topology. This is where the image comes back in.

Inferring shape is ill-posed, even if the point cloud was generated with a lidar or structured light sensor. Filling the gaps relies on assumptions about the environment. Rather than designing ad-hoc priors, we wish to use the image to inform and restrict the set of possible scenes that are compatible with the given sparse points.  

\subsubsection*{Summary of contributions}

We use a predictive cross-modal criterion to score dense depth from images and sparse depth. This kind of approach is sometimes referred to as ``self-supervised.'' Specifically, our method (i) exploits a set of constraints from temporal consistency (a.k.a. photometric consistency across temporally adjacent frames) to pose (forward-backward) consistency in a combination that has not been previously explored. To enable our pose consistency term, we introduce (ii) a set of logarithmic and exponential mapping layers for our network to represent motion using exponential coordinates, which we found to improve reconstruction compared to other parametrizations.

The challenge in using sparse depth as a supervisory (feedback) signal is precisely that it is sparse. Information at the points does not propagate to fill the domain where depth is defined. Some computational mechanism to ``diffuse the information'' from the sparse points to their neighbors is needed. Our approach proposes (iii) a simple method akin to using a piecewise planar {\em ``scaffolding''} of the scene, sufficient to transfer the supervisory signal from sparse points to their neighbors. This yields a two-stage approach, where the sparse points are first processed to design the scaffolding (``meshing and interpolation'') and then ``refined'' using the images as well as priors from the constraints just described.

One additional contribution of our approach is (iv) to introduce the first visual-inertial + depth dataset. The role of inertials is to enable reconstruction in {\em metric} scale, which is critical for robotic applications. Although scale can be obtained via other sensors, e.g., stereo, lidar, and RGB-D, we note they are not as widely available as monocular cameras with inertials (almost every modern phone has it) and consume more power. Since inertial sensors are now ubiquitous and typically co-located with cameras in many mobile devices from phones to cars, we hope this dataset will foster additional exploration into combining the complementary strengths of visual and inertial sensors.

To evaluate our method, since no other visual-inertial + depth benchmark is available, and to facilitate comparison with similar methods, we adopt the KITTI benchmark, where a Velodyne (lidar) sensor provides sparse points with scale, unlike monocular SFM, but like visual-inertial odometry (VIO). Although the biases in lidar are different from VIO, this can be considered a baseline. Note that we only use the monocular stream of KITTI (not stereo) for fair comparison.


The result is a (v) two-stage approach of scaffolding and refining with a network that contains much fewer parameters than competing methods, yet achieves state-of-the-art performance in the ``unsupervised'' KITTI benchmark (a misnomer). The supervision in the KITTI benchmark is really fusion from separate sensory channels, combined with ad-hoc interpolation and extrapolation. It is unclear whether the benefit from having such data is outweighed by the biases it induces on the estimate, and in any case such supervision does not scale; hence, we forgo (pseudo) ground truth annotations altogether.

\vspace{-0.3em}
\section{Related Work}
\noindent{\bf Supervised Depth Completion} minimizes the discrepancy between ground truth depth and depth predicted from an RGB image and sparse depth measurements. Methods focus on network topology \cite{ma2018self, uhrig2017sparsity, yang2019dense}, optimization \cite{chodosh18deep, dimitrievski2018learning, zhang2018deep}, and modeling \cite{eldesokey2018propagating, huang2019hms}. To handle sparse depth, \cite{ma2018self} employed early fusion, where the image and sparse depth are convolved separately and the results concatenated as the input to a ResNet encoder. \cite{jaritz2018sparse} proposed late fusion via a U-net containing two NASNet encoders for image and sparse depth and jointly learned depth and semantic segmentation, whereas \cite{yang2019dense} used ResNet encoders for late fusion.  \cite{eldesokey2018propagating} proposed a normalized convolutional layer to propagate sparse depth and used a binary validity map as a confidence measure. \cite{huang2019hms} proposed an upsampling layer and joint concatenation and convolution to deal with sparse inputs. All these methods require per-pixel ground-truth annotation. What is called  ``ground truth'' in the benchmarks is actually the result of data processing and aggregation of many consecutive frames. We skip such supervision and just infer dense depth by learning the cross-modal fusion from the virtually infinite volume of un-annotated data.

\noindent{\bf Unsupervised Depth Completion} methods, such as \cite{ma2018self, shivakumar2019dfusenet, yang2019dense} predict depth by minimizing the discrepancy between prediction and sparse depth input as well as the photometric error between the input image and its reconstruction from other viewpoints available only during training. \cite{ma2018self} used Perspective-n-Point (PnP) \cite{lepetit2009epnp} and Random Sample Consensus (RANSAC) \cite{fischler1981random} to align monocular image sequences for their photometric term with a second-order smoothness prior. Yet, \cite{ma2018self} does not generalize well to indoor scenes that contains many textureless regions (e.g. walls), where PnP with RANSAC may fail. \cite{shivakumar2019dfusenet} used a local smoothness term, but instead minimized the photometric error between rectified stereo-pairs where pose is known. \cite{yang2019dense} also leveraged stereo pairs and a more sophisticated photometric loss \cite{wang2004image}. \cite{yang2019dense} replaced the generic smoothness term with a\change{learned}prior to\change{regularize}their prediction.
\change{To accomplish this, \cite{yang2019dense} requires a conditional prior network (CPN) that is trained on an \textit{additional} dataset (representative of the depth completion dataset) \textit{in a fully-supervised manner} using ground-truth depth. The CPN does not generalize well outside its training domain (e.g. one cannot use a CPN trained on outdoors scenes to regularize depth predictions for indoors). Hence, \cite{yang2019dense} is essentially \textit{not} unsupervised and has limited applicability.}
In contrast, our method is\change{trained on monocular sequences, is}{\em fully unsupervised} and does not use any auxiliary ground-truth supervision.\change{Unlike previous works, our method does not require large networks (\cite{jaritz2018sparse, ma2018self, shivakumar2019dfusenet, yang2019dense}) nor any complex network operations~(\cite{eldesokey2018propagating, huang2019hms}).}Moreover, our method outperforms \cite{ma2018self, yang2019dense} on the unsupervised KITTI depth completion benchmark \cite{uhrig2017sparsity} while using fewer parameters.

\noindent{\bf Rotation Parameterization.}
To construct the photometric consistency loss during training, an auxiliary pose network is needed if no camera poses are available. While the translational part of the relative pose can be modeled as $T\in \real^3$, the rotational part belongs to the special orthogonal group $R \in SO(3) \doteq \{R\in\real^{3\times3}| R^\top R = I, \mathrm{det}(R)=+1\}$~\cite{ma2012invitation}, which is represented by a $3\times 3$ matrix. \cite{kendall2015posenet} uses quaternions, which require an {\em additional} norm constraint; this is a soft constraint imposed in the loss function, and thus is not guaranteed. \cite{fei2019geo,yin2018geonet,zhou2017unsupervised} use 
Euler angles which may result in a non-orthogonal rotation matrix due to rounding error from the multiplication of many sine and cosine terms. We use the exponential map on $SO(3)$ to map the output of the pose network to a rotation matrix. Though theoretically similar, we empirically found that the exponential map is more beneficial than the Euler angles in \secref{sec:experiments_results}. 

Our contributions are  a simple, yet effective two-stage approach resulting in a large reduction in network parameters while achieving state-of-the-art performance on the unsupervised KITTI depth completion benchmark;  using exponential parameterization of rotation for our pose network;  a pose consistency term that enforces forward and backward motion to be the inverse of each other; and finally a new depth completion benchmark for visual-inertial odometry systems with indoor and outdoor scenes and challenging motion.

\begin{figure}[]
    \centering
    \includegraphics[width=0.99\linewidth]{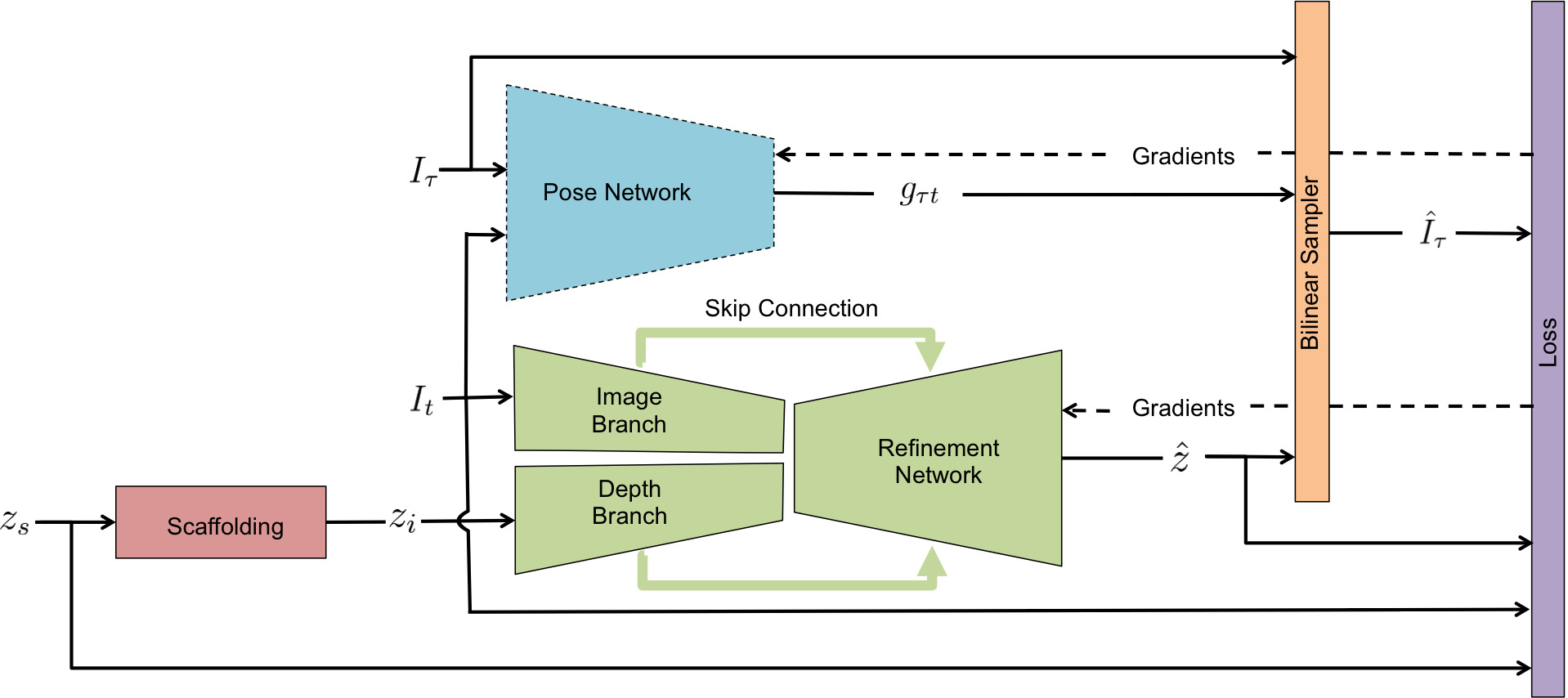}
    \vspace{-2em}
    \caption{\change{\textit{System diagram}. (best viewed in color at $5\times$). We first build a scaffolding $z_i$ from sparse depth $z_s$ estimated by VIO. Then together with the image $I_t$, $z_i$ is fed to the refinement network as input to produce output $\hat{z}$. Note: the pose network (blue) is only needed in one operation mode and is only used in training. In the other operation mode, VIO poses are used instead. The scaffolding module (red) is parameter-free -- leading to our light-weight two-stage approach.}}
    \label{fig:system-diagram}
    \vspace{-1.5em}
\end{figure}

\vspace{-0.3em}
\section{Method Formulation}
We reconstruct a 3D scene given an RGB image $I_t : \real^2 \supset \Omega \mapsto \real^3_+$ and the associated set of sparse depth measurements $z_{s} : \Omega \supset \Omega_{s} \mapsto \real_+$. 

We begin by assuming that world surfaces are graphs of smooth functions (charts) locally supported on a piecewise planar domain (scaffolding). We construct the scaffolding from the sparse point cloud (``Scaffolding'' in \figref{fig:train_progression}) to obtain $z_{i}$\change{$: \Omega \mapsto \real_+$},  then learn a completion model refining $z_{i}$ by leveraging the monocular sequences ($I_{t-1}, I_{t}, I_{t+1}$), of frames before and after the given time $t$, and the sparse depth $z_{s}$. We compose a surrogate loss $\mathcal{L}$ (\eqnref{eqn:loss_function}) for driving the training process, using an encoder-decoder architecture $f_{\theta}(\cdot)$ parameterized by weights $\theta$, where the input is an image with its scaffolding $(I_{t}, z_{i})$, and the output is the dense depth $\hat{z} = f_{\theta}(I_{t}, z_{i})$.\change{\figref{fig:system-diagram} shows an overview of our approach.}

\vspace{-0.3em}
\subsection{A Two-Stage Approach}
\vspace{-0.1em}
Depth completion is a challenging problem due to the sparsity level of the depth input, $z_{s}$. As the density of sparse depth measurements covers $\approx 5\%$ of the image plane for the outdoor self-driving scenario (\secref{sec:kitti_dataset}) and less than $\approx 1\%$ for the indoor setting (\secref{sec:void_benchmark}), generally only a single measurement will be present within a local neighborhood and in most instances none. This renders \textit{conventional convolutions ineffective} as each sparse depth measurement can be seen as a Dirac delta and convolving a kernel over the entire sparse depth input will give mostly zero activations. Hence, \cite{eldesokey2018propagating}, \cite{huang2019hms}, and \cite{uhrig2017sparsity} proposed specialized operations to propagate the information from the sparse depth input through the network.  We, instead, propose a two-stage approach that circumvents this problem by first approximating a coarse scene geometry with scaffolding and training a network to refine the approximation.

\vspace{-0.3em}
\subsection{Scaffolding}
\vspace{-0.1em}
\label{sec:scaffolding}
Given sparse depth measurements $z_{s}$, our goal is to create a coarse approximation of the scene; yet, the topology of the scene is not informed by $z_{s}$. Hence, we must rely on a prior or an assumption -- that surfaces are locally smooth and piecewise planar. We begin by applying the lifting transform~\cite{brown1979voronoi} to $z_{s}$, mapping $z_{s}$ from 2-d to 3-d space. We then compute its convex hull~\cite{barber1996quickhull}, of which the lower envelope is taken as the Delaunay triangulation of the points in $z_{s}$ -- resulting in a triangular mesh in Barycentric coordinates. 

To form the tessellation of the triangular mesh, we approximate each surface using linear interpolation within the Barycentric coordinates and the resulting scaffolding is projected back onto the image plane to produce $z_{i}$. For a given triangle, simple interpolation is sufficient for recovering the plane as a linear combination of the co-planar points. For sets of points not co-planar, interpolation will give an approximation, with which we refine using a network. We note that our approximation cannot be achieved by simply filtering (e.g. Gaussian) $z_{s}$ to propagate depth values as the filter would produce mostly zeros and even destroy the sparse depth signal.

\vspace{-0.3em}
\subsection{Refinement}
\vspace{-0.1em}
Given an RGB image and its corresponding piece-wise planar scaffolding $(I_{t}, z_{i})$, we train a network to recover the 3-d scene by refining $z_{i}$ based on information from $I_{t}$. Our network learns to refine \textit{without} ground-truth supervision by minimizing \eqnref{eqn:loss_function} (see \secref{sec:loss_function}).

\noindent\textbf{Network Architecture.}
\label{sec:network_architecture}
We propose two encoder-decoder architectures with skip connections following the late fusion paradigm \cite{jaritz2018sparse, yang2019dense}. Each encoder has an image branch and a depth branch,\change{where each contains $75\%$ and $25\%$ of the total encoder parameters, respectively.}The latent representation of the branches are concatenated and fed to the decoder. We propose a VGG11 encoder ($\approx 5.7$M parameters) containing 8 convolution layers for each branch as our best performing model, and a VGG8 encoder ($\approx 2.4$M parameters) containing only 5 convolution layers for each branch as our light-weight model. 
This is in contrast to other unsupervised methods \cite{ma2018self} (early fusion) 
and \cite{yang2019dense} (late fusion) -- both of which use ResNet34 encoders with $\approx 23.8$M and $\approx 14.8$M parameters, respectively. 
\change{\cite{ma2018self, yang2019dense} and our approach share the same decoder architecture containing $\approx 4$M parameters.}We show in \secref{sec:experiments_results} that despite having $76.1\%$ and $61.5\%$ fewer encoder parameters than \cite{ma2018self} and \cite{yang2019dense}, our VGG11 model outperforms both \cite{ma2018self} and \cite{yang2019dense}. Moreover, performance does not degrade by much from  VGG11 to VGG8 and VGG8 still surpasses \cite{ma2018self} and \cite{yang2019dense} while having a $89.9\%$ and $83.9\%$ reduction in the encoder parameters.\change{Unlike \cite{ma2018self, yang2019dense},  which requires high energy consumption hardware, 
our approach is computationally cheap, and can be deployed to low-powered agents using an Nvidia Jetson.}

\begin{figure*}[t]
\begin{adjustwidth}{-.0in}{.0in} 
\begin{center}
\includegraphics[width=1.000\textwidth]{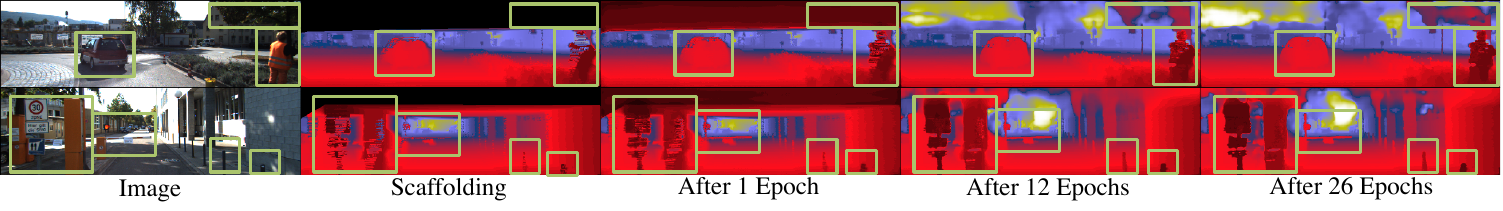}
\vspace{-2.3em}
\caption{{\em Learning to refine} (best viewed at $5\times$ with color). Our network learns to refine the input scaffolding. Green rectangles highlight the regions for comparison throughout the course of training. The network first learns to copy the input and later learns to fuse information from RGB image to refine the approximated depth from scaffolding (see row 1 pedestrian and row 2 street signs).}
\label{fig:train_progression}
\end{center}
\end{adjustwidth}
\vspace{-1.5em}
\end{figure*}

\noindent\textbf{Logarithmic and Exponential Map Layers.}
To construct our objective  (\eqnref{eqn:loss_function}), we leverage a pose network \cite{kendall2015posenet} to regress the relative camera poses $g = (R, T) \in SE(3) \doteq\{(R, T)| R\in SO(3), T\in \real^3\}$. We present a novel logarithmic map layer: $\log: SO(3) \mapsto so(3)$, where $so(3)$ is the tangent space of $SO(3)$, and an exponential map layer: $\exp:~so(3) \mapsto SO(3)$ -- for mapping $R$  between $SO(3)$ and $so(3)$.
We use the logarithmic map to construct the pose consistency loss (\eqnref{eqn:pose_consistency_loss}), and the exponential  to map the output of the pose network $\omega \doteq [\omega_1, \omega_2, \omega_3]^\top \in\real^3$ as coordinates in $so(3)$ to a rotation matrix:
\begin{equation}
R(\omega)=\exp(\hat{\omega}) \doteq I + \hat{\omega} \sin\| \omega \|_2 + \hat{\omega}^2 (1 - \cos \| \omega \|_2 )
\end{equation}
where the hat operator $\hat{\cdot}$ maps $\omega \in \real^3$ to a skew-symmetric matrix~\cite{ma2012invitation}. 
We train of our pose network using a surrogate loss (\eqnref{eqn:image_reconstruction}) without explicit supervision.\change{Ablation studies on the use of exponential coordinates and pose consistency for depth completion can be found in \tabref{tab:results_kitti_ablation} and \ref{tab:results_void_ablation}.}

Our approach contains two stages: (i) we generate a coarse piecewise planar approximation of the scene from the sparse depth inputs $z_{s}$ via scaffolding and (ii) we feed the resulting depth map along with the associated RGB image to our network for refinement (\figref{fig:train_progression}). This approach \textit{alleviates} the network from the need of learning from sparse inputs, for which \cite{ma2018self} and \cite{yang2019dense} compensated with parameters. We show the effectiveness of this approach by achieving the state-of-the-art on the unsupervised KITTI depth completion benchmark with half as many parameters as the prior-art.

\vspace{-0.3em}
\section{Loss Function}
\label{sec:loss_function}

Our loss function is a linear combination of four terms that constrain (i) the photometric consistency between the observed image and its reconstructions from the monocular sequence, (ii) the predicted depth to be similar to that of the associated available sparse depth, (iii) the composition of the predicted forward and backward relative poses to be the identity, and (iv) the prediction to adhere to local smoothness.
\begin{align}
\label{eqn:loss_function}
    \mathcal{L} = w_{ph}L_{ph}+w_{sz}L_{sz}+w_{pc}L_{pc}+w_{sm}L_{sm}
\end{align}
where $L_{ph}$ denotes photometric consistency, $L_{sz}$ sparse depth consistency, $L_{pc}$ pose consistency, and $L_{sm}$ local smoothness. Each loss term $L$ is described in the next subsections and the associated weight $w$ in \secref{sec:implementation_details}. 

\vspace{-0.3em}
\subsection{Photometric Consistency}
\vspace{-0.1em}
\label{sec:photometric_consistency_loss}
We enforce temporal consistency by minimizing the discrepancy between each observed image $I_t$ and its reconstruction  $\hat{I}_\tau$ from temporally adjacent images  $I_\tau$, where $\tau \in T \doteq \{t-1, t+1\}$:
\begin{equation}
\label{eqn:image_reconstruction}
\hat I_\tau(x) = I_\tau \big( \pi  g_{\tau t} {K}^{-1} \bar x \hat{z}(x) \big)
\end{equation}
where $\bar x = [x^T \ 1]^T$ are the homogeneous coordinates of $x\in \Omega$
, $g_{\tau t} \in SE(3)$ is the relative pose of the camera from time $t$ to $\tau$, ${K}$ denotes the camera intrinsics, and $\pi$ refers to the perspective projection. 

Our photometric consistency term is a combination of the average per pixel reprojection residual with an $L_1$ penalty and \texttt{SSIM} \cite{wang2004image}, a perceptual metric that is invariant to local illumination changes:
\begin{equation}
\begin{aligned}
\label{eqn:photometric_consistency_loss}
  	L_{ph} = \frac{1}{|\Omega|} \sum_{\tau\in T} \sum_{x \in \Omega}  
  	    &w_{co}| I_{t}(x)-\hat{I}_{\tau}(x)| + \\
  	    &w_{st}\big(1 - \texttt{SSIM}(I_t(x), \hat{I}_{\tau}(x))\big)
\end{aligned}
\end{equation}
We use $3 \times 3$ image patches centered at location $x$ for \texttt{SSIM}. $w_{co}$ and $w_{st}$ can be found in \secref{sec:implementation_details}.



\vspace{-0.3em}
\subsection{Sparse Depth Consistency}
\vspace{-0.1em}
Our sparse depth consistency term provides our predictions with {\em metric} scale by encouraging the predictions $\hat{z}$ to be similar to that of the {\em metric} sparse depth $z_{s}$ available from lidar in KITTI dataset (\secref{sec:kitti_dataset}) and sparse reconstruction in our visual-inertial dataset (\secref{sec:void_dataset}). Our sparse depth consistency loss is the $L_1$-norm of the difference between the predicted depth $\hat{z}$ and the sparse depth $z_{s}$ averaged over $\Omega_{s}$ (the support of the sparse depth):
\begin{equation}
\begin{aligned}
\label{eqn:sparse_depth_consistency_loss}
  	L_{sz} = \frac{1}{|\Omega_s|}
  	    \sum_{x \in \Omega_s}  
  	        | \hat{z}(x) - z_s(x)| 
\end{aligned}
\end{equation}

\vspace{-1em}
\subsection{Pose Consistency}
\vspace{-0.1em}
A pose network takes an ordered pair of images $(I_t, I_\tau)$ and outputs the relative pose $g_{\tau t} \in SE(3)$ (forward pose). When a temporally swapped pair $(I_\tau, I_{t})$ is fed to the network, the network is expected to output $g_{t\tau}$ (backward pose) -- the inverse of $g_{\tau t}$, \textit{i.e.}, $g_{\tau t} \cdot g_{t\tau} = e \in SE(3)$. The forward-backward pose consistency thus penalizes the deviation of the composed pose from the identity:
\begin{equation}
\label{eqn:pose_consistency_loss}
L_{pc} = \| \log(g_{\tau t} \cdot g_{t\tau}) \|_2^2
\end{equation}
where $\log: SE(3)\mapsto se(3)$ is the logarithmic map. 

\vspace{-0.3em}
\subsection{Local Smoothness}
\vspace{-0.1em}
We impose a smoothness loss on the predicted depth $\hat{z}$ by applying an $L_1$ penalty to the gradients in both the x and y directions of the predicted depth $\hat{z}$:
\begin{equation}
\begin{aligned}  
\label{eqn:smoothness_loss_depth}
  	L_{sm} = \frac{1}{|\Omega|}
  	    \sum_{x \in \Omega} 
  	    \lambda_{X}(x)|\partial_{X}\hat{z}(x)|+
  	    \lambda_{Y}(x)|\partial_{Y}\hat{z}(x)|
\end{aligned}
\end{equation}
where $\lambda_{X} = e^{-|\partial_{X}I_{t}(x)|}$ and $\lambda_{Y} = e^{-|\partial_{Y}I_{t}(x)|}$ are the edge-awareness weights to allow for discontinuities in regions corresponding to object boundaries.

\begin{figure*}[!h]
    \centering
    \includegraphics[width=1.00\textwidth]{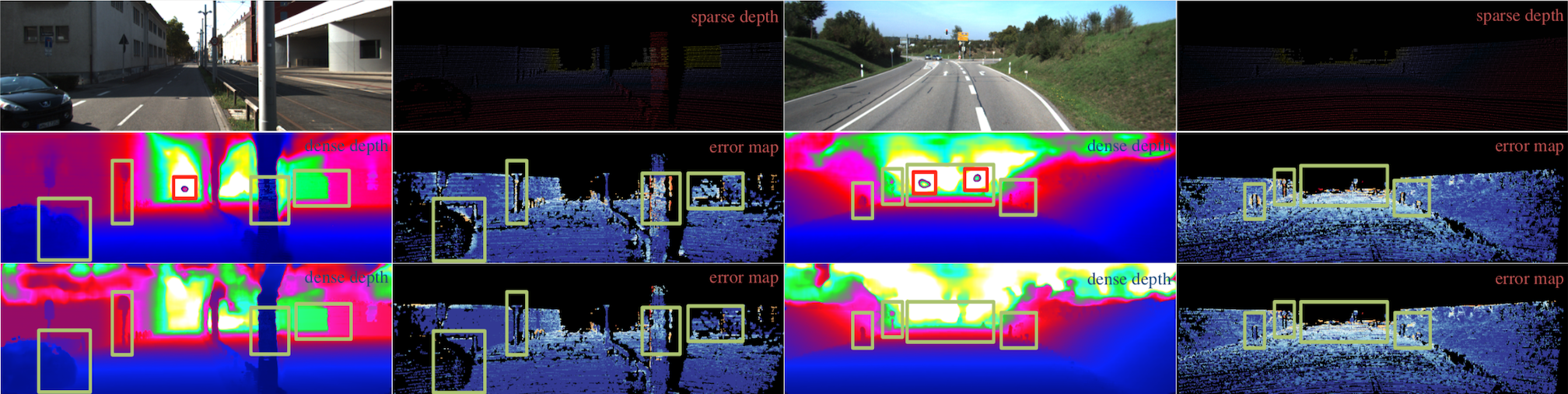}
    \vspace{-1.5em}
    \caption{\textit{Qualitative evaluation on KITTI  benchmark.} Top to bottom: input image and sparse depth, results of \cite{ma2018self}, our results. Results are taken from KITTI online test server. Warmer colors in the error map denote higher error. Green rectangles highlight regions for detail comparison. We perform better in general, particularly on thin structures and far regions. \cite{ma2018self} exhibit artifacts resembling scanlines and ``circles'' for far away regions (highlighted in red).}
    \label{fig:results_kitti_benchmark}
    \vspace{-1.5em}
\end{figure*}

\vspace{-0.3em}
\subsection{The Role of Inertials}
\vspace{-0.1em}
Although inertials are not directly present in the loss, their role in \textit{metric} depth completion is crucial. Without inertials, a SLAM system cannot produce sparse point clouds in metric scale, which are then used as both the input to the scaffolding stage (\secref{sec:scaffolding}) and a supervisory signal (\eqnref{eqn:sparse_depth_consistency_loss}).

\vspace{-0.3em}
\section{Datasets}
\vspace{-0.1em}
\subsection{KITTI Benchmark}
\label{sec:kitti_dataset}
\vspace{-0.1em}
We evaluate our approach on the KITTI depth completion benchmark \cite{uhrig2017sparsity}. The dataset provides $\approx 80,000$ raw image frames and associated sparse depth maps. The sparse depth maps are the raw output from the Velodyne lidar sensor, each with a density of $\approx 5\%$. The ground-truth depth map is created by accumulating the neighbouring 11 raw lidar scans, with dense depth corresponding to the bottom $30\%$ of the images. We use the officially selected 1,000 samples for validation and we apply our method to 1,000 testing samples, with which we submit to the official KITTI website for evaluation. The results are reported in  \tabref{tab:results_kitti_benchmark}.

\vspace{-0.4em}
\subsection{VOID Benchmark}
\vspace{-0.1em}
\label{sec:void_dataset}
While KITTI provides a standard benchmark for evaluating depth completion in the driving scenario, there exists no standard depth completion benchmark for the indoor scenario. \cite{ma2018self, yang2019dense} used NYUv2 \cite{Silberman2012} -- an RGB-D dataset -- to develop and evaluate their models on indoor scenes. Yet, each performs a different evaluation protocol with different sparse depth samples -- varying densities of depth values were randomly sampled from the depth frame, preventing direct comparisons between methods. Though this is reasonable as a proof of concept, it is not realistic in the sense that no sensor measures depth at random locations. 

\begin{figure}[t]
    \centering
    \includegraphics[width=0.7\linewidth]{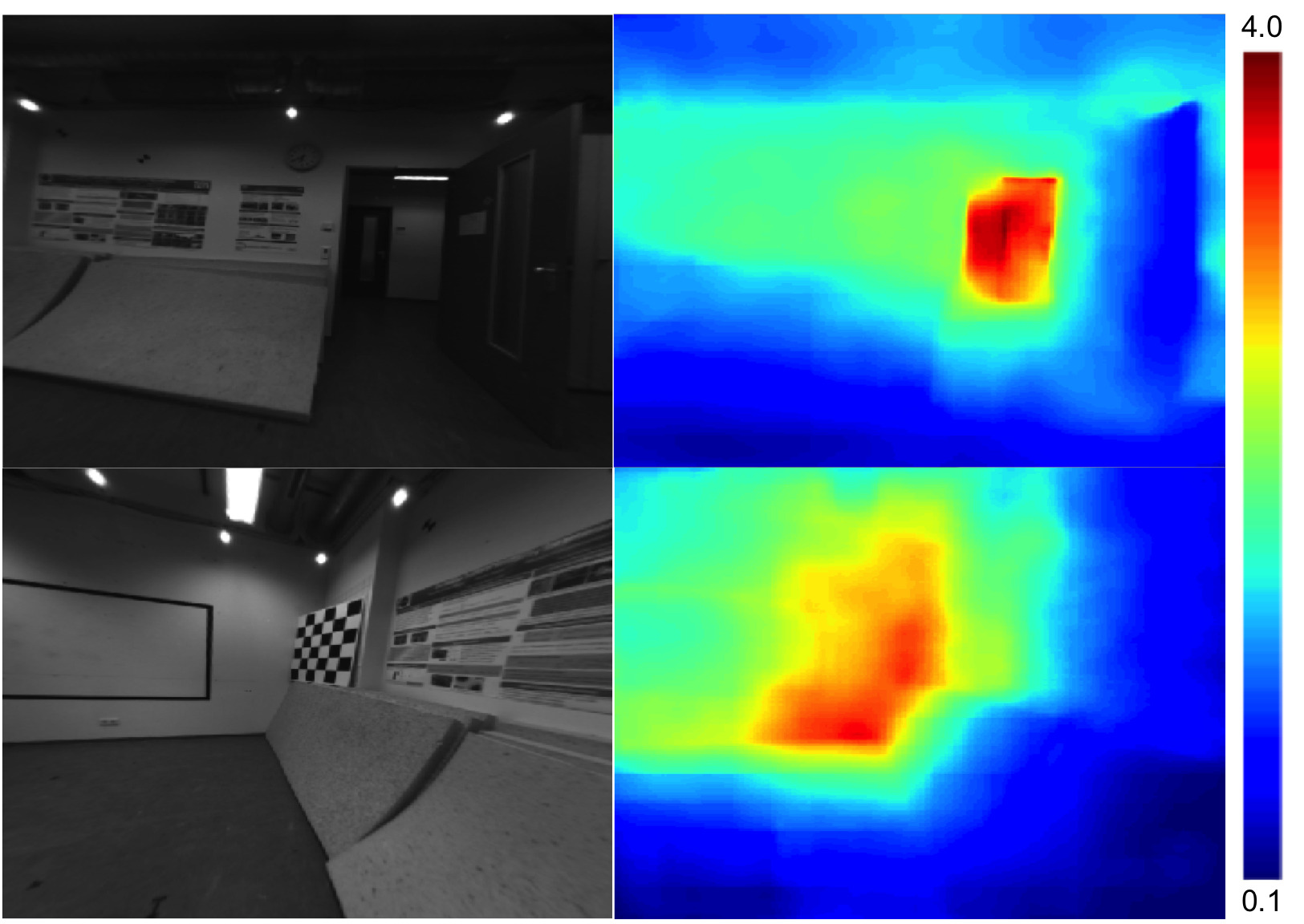}
    \vspace{-0.7em}
    \caption{\change{\textit{Qualitative results on TUM-VI} (best viewed in color at $2\times$). We apply our method to TUM-VI and obtained our results using sparse depth input at a density level of $0.015\%$. Unlike KITTI and VOID, TUM-VI images are monochrome, and bear a highly distorted fisheye camera model, which was compensated in training. Color bar shows the depth range.}}
    \vspace{-0.4em}
    \label{fig:tumvi-examples}
\end{figure}

\noindent\textbf{The VOID dataset.} We propose a new publicly available dataset for a real world use case of depth completion by bootstrapping sparse reconstruction in \textit{metric} space from a SLAM system. While it is well known that metric scale is not observable in the purely image-based SLAM and SFM setting, it has been resolved by the recent advances in VIO \cite{joness10ijrr, mourikis2007multi}, where metric pose and structure estimation can be realized in a gravity-aligned and scaled reference frame using a inertial measurement unit (IMU). 
To this end, we leverage an off-the-shelf VIO system~\footnote{\url{https://github.com/ucla-vision/xivo}}, atop which we construct our dataset and develop our depth completion model. While there are some visual-inertial datasets (e.g. TUM-VI \cite{schubert2018tum} and PennCOSYVIO \cite{pfrommersdc17}), they lack per-frame dense depth measurements for cross-modal validation, and are also relatively small -- rendering them unsuitable for training deep learning models.\change{To demonstrate the applicability of our approach, we additionally show qualitative results on the TUM-VI dataset in \figref{fig:tumvi-examples} using sparse depth density level of $0.015\%$.}

Our dataset is dubbed ``Visual Odometry with Inertial and Depth'' or ``VOID'' for short and is comprised of RGB video streams and inertial measurements for \textit{metric} reconstruction along with per-frame dense depth for cross-modal validation.  

\begin{figure}[t!]
    \centering
    \includegraphics[width=0.89\linewidth]{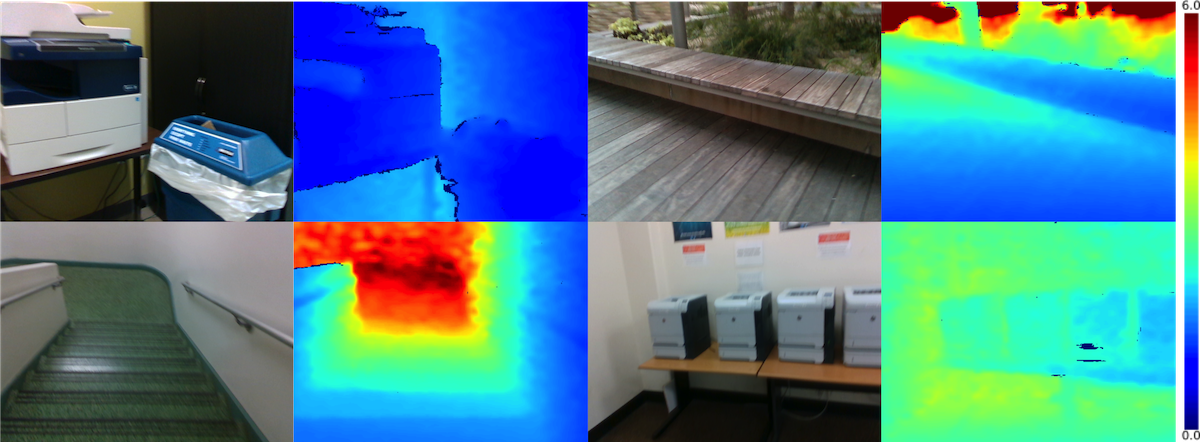}
    \vspace{-1em}
    \caption{\textit{Sample RGB + D images} in the VOID dataset (best viewed in color at $5\times$). Color bar shows the depth range.}
    \label{fig:void-examples}
    \vspace{-2em}
\end{figure}

\noindent\textbf{Data acquisition.} Our data was collected using the latest Intel RealSense D435i camera~\footnote{\url{https://realsense.intel.com/depth-camera/}}, which was configured to produce synchronized accelerometer and gyroscope measurements at 400 Hz, along with synchronized VGA-size ($640 \times 480$) RGB and depth streams at 30 Hz. The depth frames are acquired using active stereo and is aligned to the RGB frame using the sensor factory calibration (see \figref{fig:void-examples}). All the measurements are time-stamped.

The SLAM system we use is based on \cite{joness10ijrr} -- an EKF-based VIO model. While the VIO recursively estimates a joint posterior of the state of the sensor platform (e.g. pose, velocity, sensor biases, and camera-to-IMU alignment) and a small set of reliable feature points, the 3D structure it estimates is extremely sparse --  typically $20 \sim 30$ feature points (in-state features). To facilitate 3D reconstruction, we track a moderate amount of out-of-state features in addition to the in-state ones, and estimate the depth of the feature points using auxiliary depth sub-filters~\cite{ma2012invitation}.

\noindent\textbf{The benchmark.} We evaluate our method on the VOID depth completion benchmark, which contains 56 sequences in total, both indoor and outdoor with challenging motion. Typical scenes include classrooms, offices, stairwells, laboratories, and gardens. Of the 56 sequences, 48 sequences ($\sim 40K$ frames) are designated for training and 8 sequences for testing, from which we sampled $800$ frames to construct the testing set. 
Our benchmark provides sparse depth maps at three density levels. We configured our SLAM system to track and estimate depth of 1500, 500 and 150 feature points, corresponding to $0.5\%, 0.15\%$ and $0.05\%$ density of VGA size, which are then used in the depth completion task. 

\vspace{-0.3em}
\section{Implementation Details}
\label{sec:implementation_details}
Our approach was implemented using TensorFlow \cite{abadi2016tensorflow}. With a Nvidia GTX 1080Ti, training takes $\approx 42$ hours for our VGG11 model and $\approx 34$ hours for our VGG8 model on KITTI depth completion benchmark (\secref{sec:kitti_dataset}) for 30 epochs; whereas training takes $\approx 10$ hours and $\approx 7$ hours on the VOID benchmark (\secref{sec:void_dataset}) for 10 epochs. Inference takes $\approx$ 22 ms per image. We used Adam \cite{kingma2014adam} with $\beta_1=0.9$ and $\beta_2=0.999$ to optimize our network end-to-end with a base learning rates of $1.2 \times 10^{-4}$ for KITTI and $1 \times 10^{-4}$ for \void. We decrease the learning rate by half after 18 epochs for KITTI and 6 epochs for \void, and again after 24 epochs and 8 epochs, respectively. We train our network with a batch size of 8 using a $768 \times 320$ resolution for KITTI and $640 \times 480$ for \void. We are able to achieve our results on the KITTI benchmark using the following set of weights for each term in our loss function: $w_{ph}=1.00$, $w_{co}=0.20$, $w_{st}=0.40$, $w_{sz}=0.20$, $w_{pc}=0.10$ and $w_{sm}=0.01$. For the VOID benchmark, we increased $w_{sz}$ to $1.00$ and $w_{sm}$ to $0.10$. We do not use any data augmentation.  

\begin{table}
\caption{Error metrics.}
\vspace{-1em}
\begin{adjustwidth}{-.0in}{-.0in} 
\footnotesize
\centering
\setlength\tabcolsep{12pt}
\begin{tabular}{ccl}
\toprule
    Metric & units & Definition \\ \midrule
    MAE & {\it mm} &$\frac{1}{|\Omega|} \sum_{x\in\Omega} |\hat z(x) - z_{gt}(x)|$ \\
    RMSE & {\it mm} & $\big(\frac{1}{|\Omega|}\sum_{x\in\Omega}|\hat z(x) - z_{gt}(x)|^2 \big)^{1/2}$ \\
    iMAE & {\it 1/km} & $\frac{1}{|\Omega|} \sum_{x\in\Omega} |1/ \hat z(x) - 1/z_{gt}(x)|$ \\
    iRMSE & {\it 1/km} & $\big(\frac{1}{|\Omega|}\sum_{x\in\Omega}|1 / \hat z(x) - 1/z_{gt}(x)|^2\big)^{1/2}$ \\ \bottomrule
\end{tabular}
\end{adjustwidth}
\vspace{0.2em}
\begin{tablenotes}
    Error metrics for evaluating KITTI and VOID depth completion benchmarks, where $z_{gt}$ is the ground truth.
\end{tablenotes}
\vspace{-2em}
\label{tab:error_metrics}
\end{table}

\vspace{-0.3em}
\section{Experiments and Results}
\label{sec:experiments_results}
\subsection{KITTI Depth Completion Benchmark}
\label{sec:kitti_depth_completion_benchmark}
\vspace{-0.1em}
\change{We show quantitative and qualitative comparisons on the unsupervised KITTI depth completion benchmark in \tabref{tab:results_kitti_benchmark} and \figref{fig:results_kitti_benchmark}, respectively.}The results of the methods listed are taken directly from their papers. We note that \cite{yang2019dense} only reported their result in their paper and do have have an entry in KITTI depth completion benchmark for their unsupervised model. Hence, we compare qualitatively with the prior-art \cite{ma2018self}.\change{Our VGG11 model outperforms the state-of-the-art \cite{yang2019dense} on every metric by as much as $12.8\%$ while using $48.4\%$ fewer parameters.
Our light-weight VGG8 model also outperforms \cite{yang2019dense} on MAE, RMSE, and iMAE while \cite{yang2019dense} beat our VGG8 by 2.2\% on iRMSE.}We note that \cite{yang2019dense} trains a \change{separate network, using ground truth, to supervise their depth completion model.}Moreover, \cite{yang2019dense} exploits rectified stereo-imagery where the pose of the cameras is known; whereas, we learn our pose by jointly training the pose network with our depth predictor. In comparison to \cite{ma2018self} (who also uses monocular videos),\change{both our VGG11 and VGG8 model outperforms them on every metric while using much fewer paramters.}We also note that the qualitative results of \cite{ma2018self} contains artifacts such as apparent scanlines of the Velodyne and ``circles'' in far regions. 

\begin{figure}
    \centering
    \includegraphics[width=0.35\textwidth]{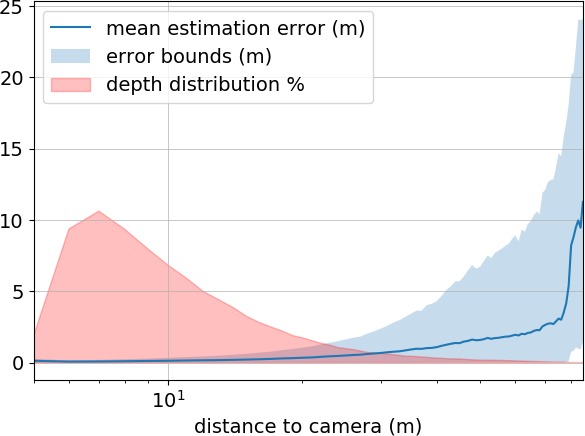}
    \vspace{-0.7em}
    \caption{\textit{Error characteristics of our model on KITTI.} The abscissa shows the distance of sparse data points measured by Velodyne, of which the percentage of all the data points is shown in red; the blue curve shows the mean absolute error of the estimated depth at the given distance, of which the 5-{\it th} and 95-{\it th} percentile enclose the light blue region.}
    \label{fig:error_vs_distance}
    \vspace{-1.5em}
\end{figure}

As an introspective exercise, we plot the mean error of our model at varying distances on the KITTI validation set (\figref{fig:error_vs_distance}) and overlay it with the ground truth depth distribution to show that our model performs very well in distances that matter in real-life scenarios. Our performance begins to degrade at distances larger than 80 meters; this is due to the lack of sparse measurements and insufficient parallax -- problems that plague methods relying on multi-view supervision. 

\begin{table}
\caption{KITTI depth completion benchmark.}
\vspace{-1em}
\begin{adjustwidth}{-.0in}{-.0in} 
\footnotesize
\centering
\setlength\tabcolsep{5.0pt}
\begin{tabular}{l c c c c c}
    \toprule
    Method &\change{\# Parameters}& MAE & RMSE & iMAE & iRMSE 
    \\ \midrule
    Schneider \cite{schneider2016semantically}     
    &\change{{\rm not reported}}& 605.47        & 2312.57       & 2.05          & 7.38 \\ 
    Ma \cite{ma2018self}
    &\change{$\approx 27.8M$}& 350.32        & 1299.85       & 1.57          & 4.07 \\ 
    Yang \cite{yang2019dense}
    &\change{$\approx 18.8M$}& 343.46        & 1263.19       & 1.32          & 3.58 \\ 
    Ours VGG11
    &\change{$\approx 9.7M$}& {\bf 299.41} & 1169.97 & {\bf 1.20} & {\bf 3.56} \\
    Ours VGG8
    &\change{$\approx 6.4M$}& 304.57 & {\bf 1164.58} & 1.28 & 3.66 \\
    \bottomrule
\end{tabular}
\end{adjustwidth}
\vspace{0.2em}
\begin{tablenotes} 
    We compare our model to unsupervised methods on the KITTI depth completion benchmark \cite{uhrig2017sparsity}.\change{Number of parameters used by each are listed for comparison. \cite{schneider2016semantically} stated that they use a fully convolution network, but does not specify the full architecture. Our VGG11 model outperforms state-of-the-art \cite{yang2019dense} across all metrics. Despite reducing $\approx 3.3$M parameters, our VGG8 model does not degrade by much and outperforms VGG11 marginally on the RMSE metric. Moreover, our VGG8 model also outperforms \cite{ma2018self} and \cite{yang2019dense}.} 
\end{tablenotes}
\label{tab:results_kitti_benchmark}
\vspace{-1.1em}
\end{table}

\begin{table}
\caption{KITTI depth completion ablation study.}
\vspace{-1em}
\footnotesize
\centering
\setlength\tabcolsep{1.5pt}
\begin{tabular}{l c c c c c c}
    \toprule
    Model & Encoder & Rot. & MAE & RMSE & iMAE & iRMSE \\ \midrule
    Scaffolding  & - & - & 443.57 & 1990.68 & 1.72 & 6.43 \\ 
    
    $L_{ph}$ + $L_{sz}$ + $L_{sm}$ (vanilla) & VGG11 & Eul.     
    & 347.14 & 1330.88 & 1.46 & 4.22 \\ 
    
    $L_{ph}$ + $L_{sz}$ + $L_{sm}$ & VGG11 & Eul.     
    & 327.84 & 1262.46 & 1.31 & 3.87 \\ 
    
    $L_{ph}$ + $L_{sz}$ + $L_{sm}$ & VGG11 & Exp.
    &  312.10 & 1255.21 & 1.28 & 3.86 \\    
    
    $L_{ph}$ + $L_{sz}$ + $L_{pc}$ + $L_{sm}$ & VGG11 & Exp.
    & {\bf 305.06} & 1239.06 & {\bf 1.21} & {\bf 3.71} \\ 
    
    $L_{ph}$ + $L_{sz}$ + $L_{pc}$ + $L_{sm}$ & VGG8 & Exp.
    & 308.81 & {\bf 1230.85} & 1.29 & 3.84 \\
    \bottomrule
\end{tabular}
\setlength{\belowcaptionskip}{-5pt}
\vspace{0.2em}
\begin{tablenotes}
    We compare variants of our model on the KITTI depth completion validation set. Each model is denoted by its loss function. Regions with missing depth in Scaffolding Only is assigned average depth. It is clear that scaffolding alone (row 1) and our baseline model trained {\em without} scaffolding (row 2) do poorly compared to our models that combine both (rows 3-6). Our full model using VGG11 produces the best overall results and achieves state-of-the-art on the test set \tabref{tab:results_kitti_benchmark}.\change{Our approach is robust, our light-weight VGG8 model achieves similar performance to our VGG11 model.}
\end{tablenotes}
\label{tab:results_kitti_ablation}
\vspace{-2em}
\end{table}

\vspace{-0.5em}
\subsection{KITTI Depth Completion Ablation Study}
\label{sec:kitti_depth_completion_ablation_study}
\vspace{-0.1em}
We analyze the effect brought by each of our contributions through a quantitative evaluation on the KITTI depth completion validation set (\tabref{tab:results_kitti_ablation}). Our two baseline models, scaffolding and vanilla model trained without scaffolding, perform poorly in comparison to the models that are trained with scaffolding -- showcasing the effectiveness of our refinement approach. Although the loss functions are identical, exponential parameterization consistently improves over Euler angles across all metrics.\change{We believe this is due to the regularity of the derivatives of the exponential map \cite{gallego2015compact} compared to other parameterizations -- resulting in faster convergence and wider minima during training.} While \cite{fei2019geo, wang2018learning, yin2018geonet} train their pose network using the photometric error 
with no additional constraint, we show that 
it is beneficial to impose our pose consistency term (\secref{eqn:pose_consistency_loss}). By constraining the forward and backward poses to be inverse of each other, we obtain a more accurate pose resulting in better depth prediction. Our experiments verify this claim as we see an improvement in\change{across all metrics}in
\tabref{tab:results_kitti_ablation}. We note that the improvement does not seem significant on KITTI as the motion is mostly planar; however, when predicting non-trivial 6 DoF motion (\secref{sec:void_ablation}), we see a significant boost when employing this term. Our model trained with the full loss function produces the best results (bolded in \tabref{tab:results_kitti_benchmark}) and is the state-of-the-art for unsupervised KITTI depth completion benchmark. We further propose a 
VGG8 model that only contains $\approx 6.4$M parameters.\change{Despite having $34\%$ fewer paramters than VGG11, the performance of VGG8 does not degrade by much 
(see \tabref{tab:results_kitti_benchmark}, \ref{tab:results_kitti_ablation}, \ref{tab:results_void_benchmark}).}

\vspace{-0.7em}
\subsection{VOID Depth Completion Benchmark}
\label{sec:void_benchmark}
\begin{figure}
\centering
\includegraphics[width=0.40\textwidth]{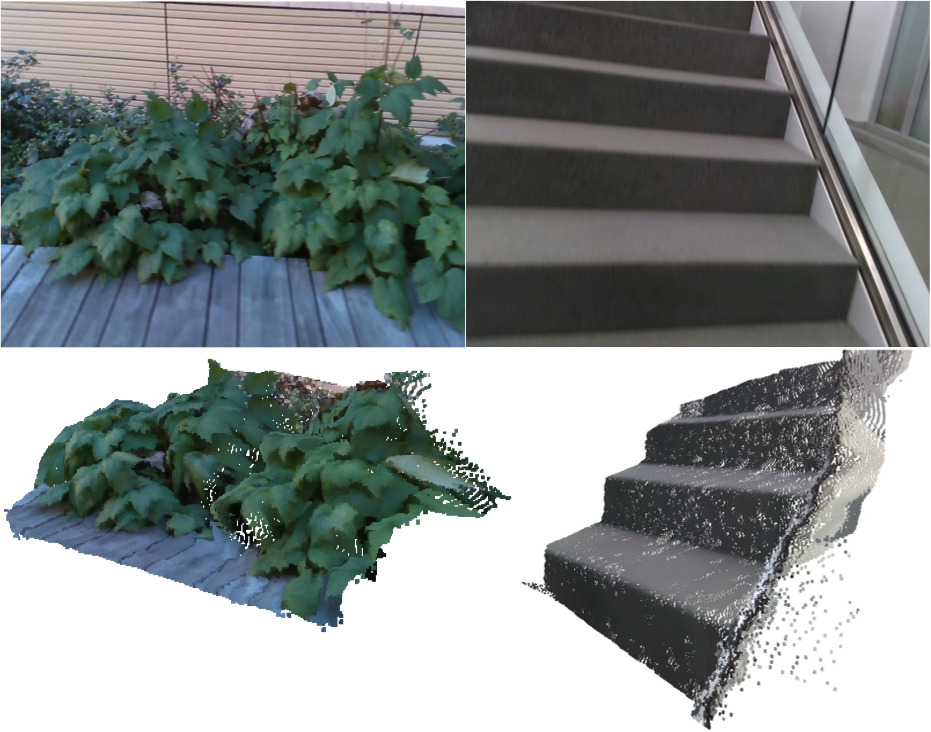} \\
\vspace{-0.9em}
\caption{\change{\textit{Qualitative evaluation on VOID benchmark.} Top: Input RGB images. Bottom: Densified depth images back-projected to 3D, colored, and viewed from a different vantage point.}}
\label{fig:results_void_benchmark}
\vspace{-1.5em}
\end{figure}

\vspace{-0.1em}
We evaluate our method on the VOID depth completion benchmark for all three density levels (\tabref{tab:results_void_benchmark}) using error metrics in \tabref{tab:error_metrics}. As the photometric loss (\eqnref{eqn:photometric_consistency_loss}) is largely dependent on obtaining the correct pose, we additionally propose a hybrid model, where the relative camera poses from our visual-inertial SLAM system are used to construct the photometric loss to show\change{an}upper bound on performance.   
In contrast to the KITTI, 
which provides\change{$\approx 5\%$ sparse depth 
density}concentrated on the bottom 30\% of the image, the VOID benchmark only provides $\approx 0.5\%$, $\approx 0.15\%$ and $\approx 0.05\%$ densities in sparse depth.
Yet, our method is still able to produce reasonable results for indoor scenes with a MAE of $\approx 8.5$ cm on $0.5\%$ density and $\approx 17.9$ cm when given only $0.05\%$. Since most scenes contain textureless regions, sparse depth supervision becomes important as photometric reconstruction is unreliable. Hence, 
\change{performance degrades as density decreases.}Yet, we degrade gracefully: as density decreases by 10X, our error only doubles. We note that the scaffolding may poorly represent the scene. In the worst case, where it provides no extra information, our method becomes the common depth completion approach. Also, we observe systematic performance improvement in all the evaluation metrics (\tabref{tab:results_void_benchmark}) when replacing the pose network with SLAM pose. This can be largely attributed to the necessity for the correct pose to minimize photometric error during training. Our pose network may not be able to consistently predict the correct pose due to the challenging motion of the dataset. \figref{fig:results_void_benchmark} shows two sample RGB images with the densified depth images back-projected to 3D, colored, and viewed from a different vantage point.

\setlength{\tabcolsep}{1pt}

\vspace{-0.7em}
\subsection{VOID Depth Completion Ablation Study}
\label{sec:void_ablation}
\vspace{-0.1em}
To better understand the effect of rotation parameterization and our pose consistency loss (\eqnref{eqn:pose_consistency_loss}) on the depth completion task, we compare variants of our model and again replace the pose network with SLAM pose to show an upper-bound on performance. Although exponential outperforms Euler parameterization, we note that\change{both perform much worse than using SLAM pose.} 
However, we observe a performance boost when applying our pose consistency term and our model improves over exponential without pose consistency by\change{as much as $23.4\%$. Moreover, it approaches the performance of our model trained using SLAM pose.}
This trend still holds when density decreases (\tabref{tab:results_void_benchmark}). This suggests that despite the additional constraint, the pose network still has some difficulties predicting the pose due to the challenging motion. This finding, along with results from \tabref{tab:results_void_benchmark}, highlights the strength of classical SLAM systems in the deep learning era, which also urges us to develop and test pose networks on the VOID dataset which features non-trivial 6 DoF motion -- much more challenging than the mostly-planar motion in KITTI.
\begin{table}[]
\caption{\void depth completion benchmark and ablation study.}
\vspace{-1em} 
\footnotesize
\centering
\setlength\tabcolsep{5.5pt}
\begin{tabular}{l c c c c}
    \toprule
    Method & MAE & RMSE & iMAE & iRMSE \\
    \midrule
    \change{Ma \cite{ma2018self}}
    &\change{198.76}&\change{260.67}&\change{88.07}&\change{114.96}\\
    \change{Yang \cite{yang2019dense}}
    &\change{151.86}&\change{222.36}&\change{74.59}&\change{112.36}\\
    VGG11 PoseNet + Eul. 
    & 108.97 & 212.16 & 64.54 & 142.64 \\ 
    VGG11 PoseNet + Exp.
    & 103.31 & 179.05 & 63.88 & 131.06 \\ 
    VGG11 PoseNet + Exp. + $L_{pc}$
    & 85.05 & 169.79 & 48.92 & 104.02 \\ 
    VGG11 SLAM Pose 
    & {\bf 73.14} & {\bf 146.40} & {\bf 42.55} & {\bf 93.16} \\
    \change{VGG8 PoseNet + Exp. + $L_{pc}$}
    &\change{94.33}&\change{168.92}&\change{56.01}&\change{111.54}\\ 
    \bottomrule
\end{tabular}
\setlength{\belowcaptionskip}{-5pt}
\vspace{0.2em}
\begin{tablenotes}
    We compare the variants of our pose network. SLAM Pose replaces the output of pose network with SLAM estimated pose to gauge an upper bound in performance. When using our pose consistency term with exponential parameterization, our method approaches the performance of our method when using SLAM pose.\change{Note: we trained \cite{ma2018self} from scratch using ground-truth pose and adapted [26] to train on monocular sequences. The conditional prior network used in \cite{yang2019dense} is trained on ground truth from  NYUv2 \cite{Silberman2012}.}
\end{tablenotes}
\label{tab:results_void_ablation}
\vspace{-1em}
\end{table}

\begin{table}
\caption{Depth completion on \void with varying sparse depth density.}
\vspace{-1em}
\footnotesize
\centering
\setlength\tabcolsep{7.5pt}
\begin{tabular}{l c c c c c}
    \toprule
    Density & Pose From & MAE & RMSE & iMAE & iRMSE \\ \midrule
    \multirow{2}{*}{$\sim 0.5\%$}      
    & PoseNet & 85.05 & 169.79 & 48.92 & 104.02 \\
    & SLAM & 73.14 & 146.40 & 42.55 & 93.16 \\
    \midrule
    \multirow{2}{*}{$\sim 0.15\%$}            
    & PoseNet & 124.11 & 217.43 & 66.95 & 121.23 \\
    & SLAM & 118.01 & 195.32 & 59.29 & 101.72 \\
    \midrule
    \multirow{2}{*}{$\sim 0.05\%$}      
    & PoseNet & 179.66 & 281.09 & 95.27 & 151.66 \\
    & SLAM & 174.04 & 253.14 & 87.39 & 126.30 \\
    \bottomrule
\end{tabular}
\vspace{0.2em}
\begin{tablenotes}
     The \void dataset contains VGA size images ($480\times 640$) of both indoor and outdoor scenes with challenging motion. For ``Pose From'', SLAM refers to relative poses estimated by a SLAM system, and PoseNet refers to relative poses predicted by a pose network.
\end{tablenotes}
\label{tab:results_void_benchmark}
\vspace{-2.1em}
\end{table}

\vspace{-0.7em}
\section{Discussion}

While deep networks have attracted a lot of attention as a general framework to solve an array of problems, we must note that pose may be difficult to learn on datasets with non-trivial 6 DoF motion -- which the SLAM community has studied for decades. We hope that VOID will serve as a platform to develop models that can handle challenging motion and further foster fusion of multi-sensory data. Furthermore, we show that a network can recover the scene geometry from extremely sparse point clouds (e.g. features tracked by SLAM). We also show that improvements can be obtained by leveraging pose from a SLAM system instead of a pose network. These findings motivate a possible mutually beneficial marriage between classical methods and deep learning. 

{
\vspace{-1.3em}
\bibliographystyle{IEEEtran}
\bibliography{egbib}
}

\onecolumn 

\begin{appendices}

\section{VOID Dataset}
\begin{figure}[h]
    \centering
    \begin{tabular}{cc}
    \includegraphics[height=0.25\textheight]{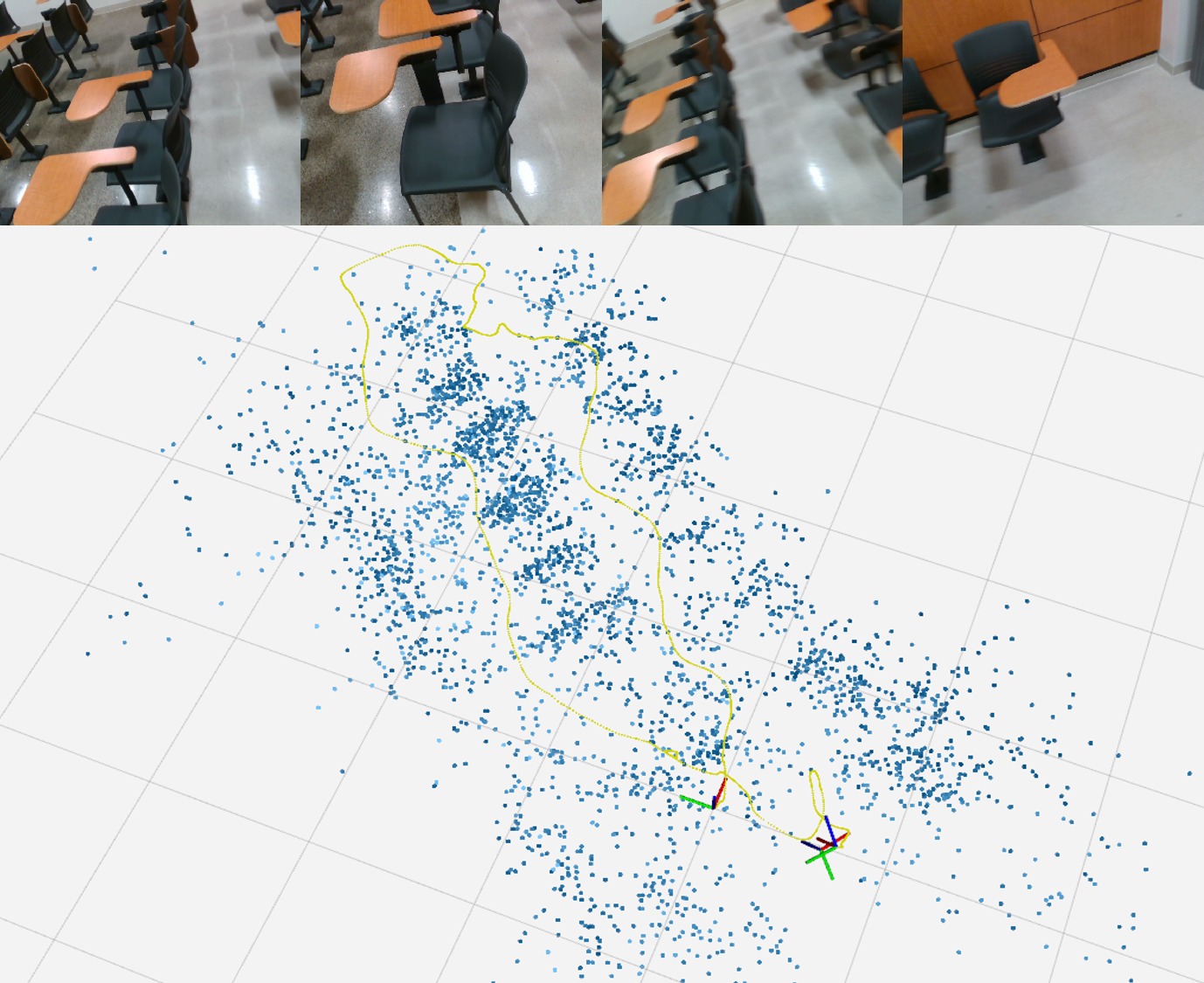} & 
    \includegraphics[height=0.25\textheight]{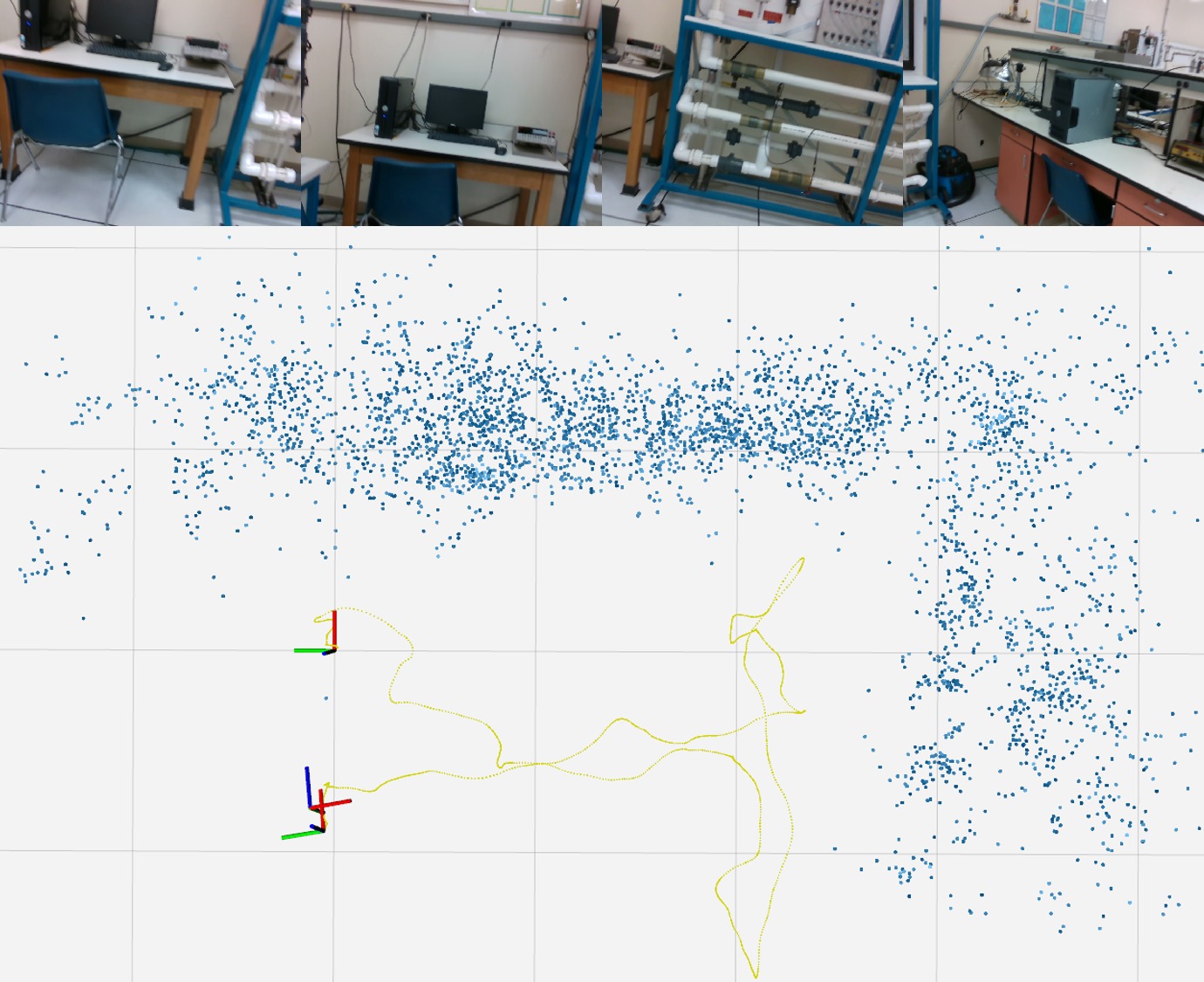} \\
    Two rows of chairs in a classroom & ``L'' shape formed by desks in a mechanical laboratory\\
    \includegraphics[height=0.25\textheight]{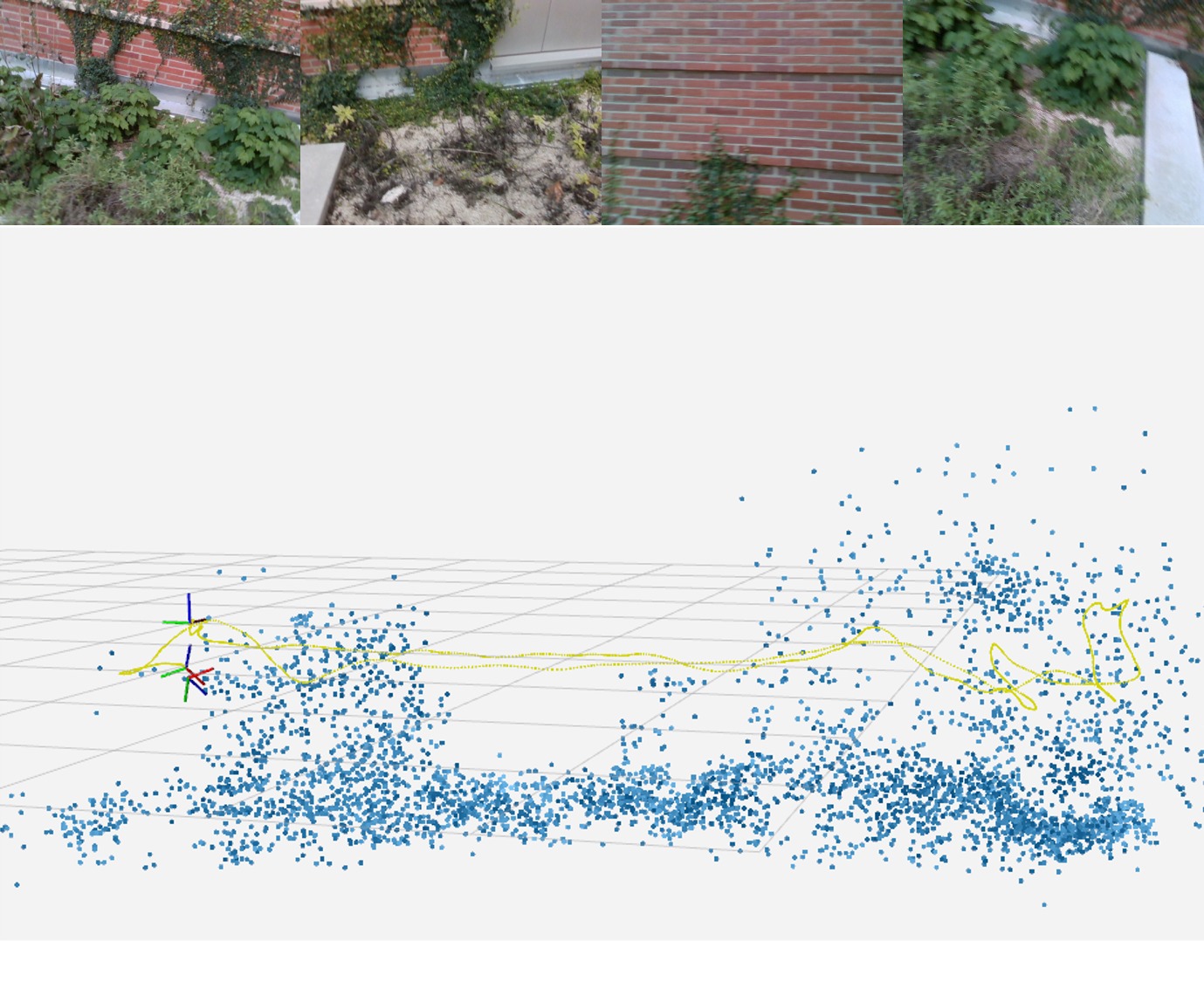} &
    \includegraphics[height=0.25\textheight]{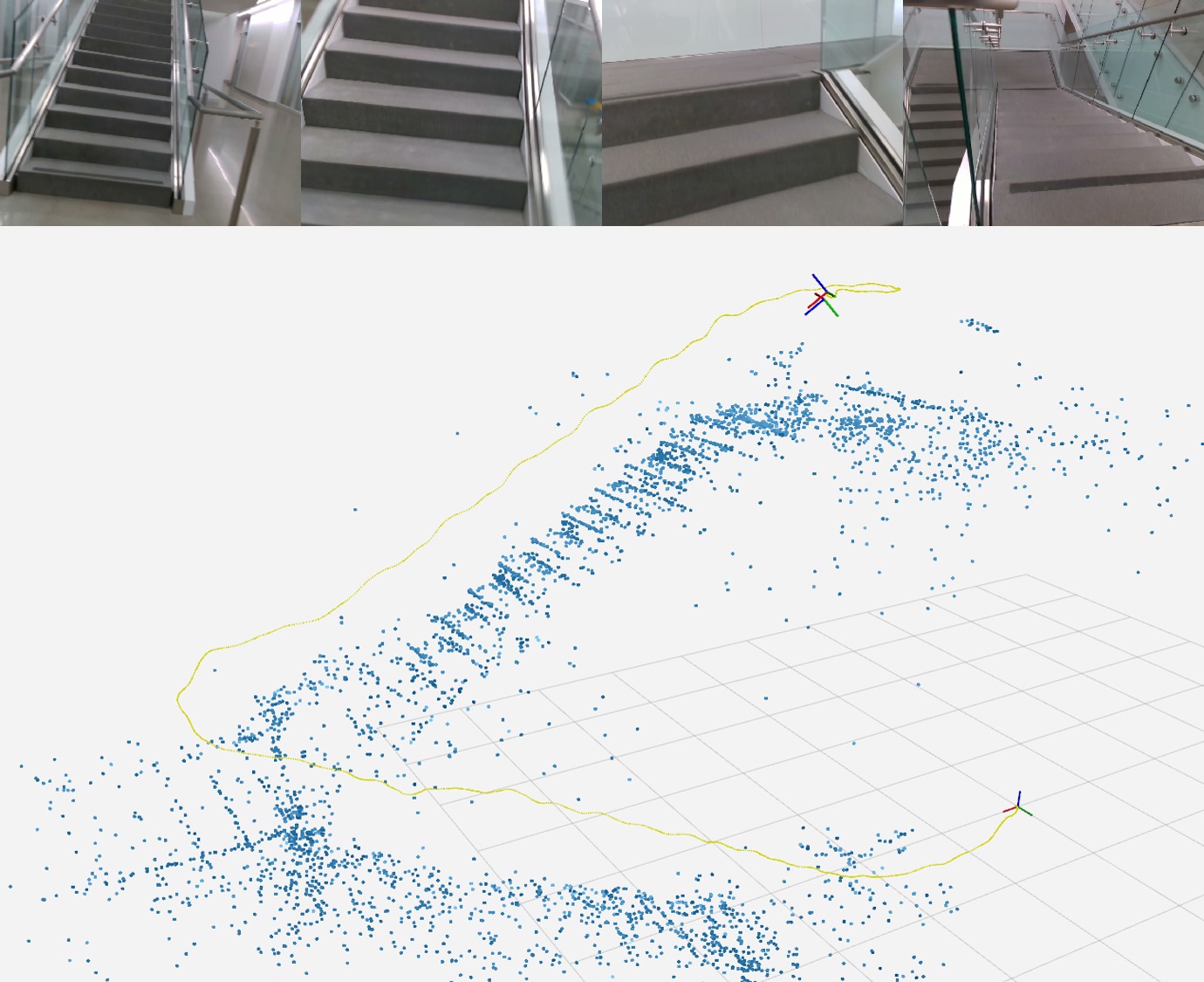} \\
    a brick wall with plants on the ground underneath & stairs
    \end{tabular}
    \caption{\textit{Sample sequences in VOID dataset} (best viewed in color at $5\times$). In each panel, the top inset shows 4 sample images of a video sequence in our VOID dataset; the bottom shows the sparse pointcloud reconstruction (blue) and camera trajectory (yellow) from our VIO.}
    \label{fig:void_dataset}
\end{figure}

In the main paper, we introduced the ``Visual Odometry with Inertial and Depth'' (VOID) dataset with which we propose a new depth completion benchmark. We described the data acquisition process, benchmark setup, and evaluation protocols in \secref{sec:void_dataset} and \secref{sec:void_benchmark}. To give some flavor of the VOID dataset, \figref{fig:void_dataset} shows a set of images (top inset) sampled from video sequences in VOID, and output of our visual-inertial odometry (VIO) system (bottom), where the blue pointcloud is the sparse reconstruction of the underlying scene and the yellow trace is the estimated camera trajectory.

\clearpage

\section{More results on VOID Dataset}
\begin{figure}[h]
\centering
\begin{tabular}{cc}
\includegraphics[height=0.45\textheight]{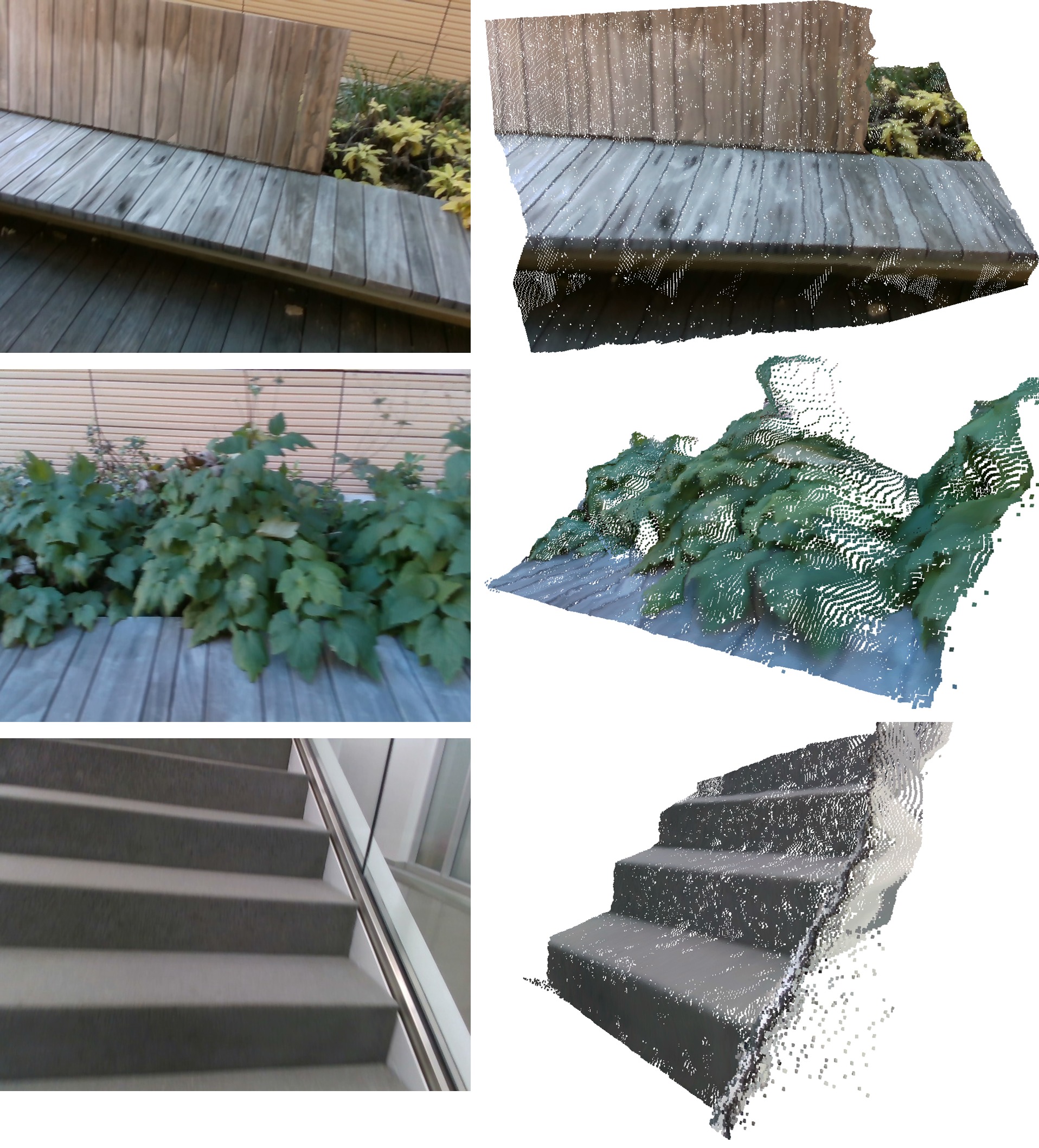} &
\includegraphics[height=0.470\textheight]{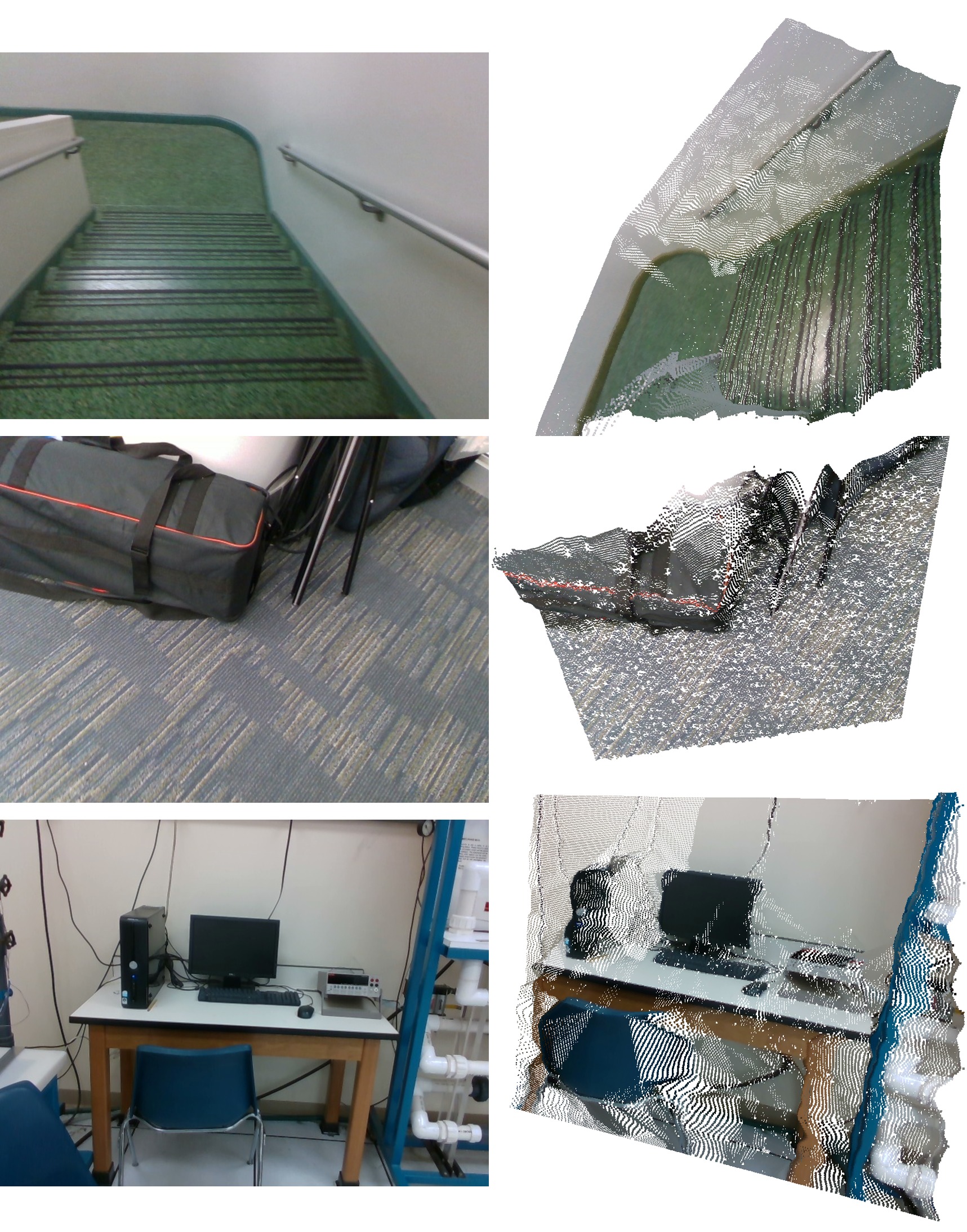}
\end{tabular}
\caption{\textit{Qualitative results on VOID dataset.} In each panel, the left shows a sample RGB image fed to our depth completion network as input; the right shows the completed depth map back-projected to 3D, colored, and viewed from a different vantage point. Our method recovers the scene structure with details at various ranges in both indoor and outdoor settings.}
\label{fig:void_additional_results}
\end{figure}

In the main paper, we evaluated our approach on the VOID depth completion benchmark in \secref{sec:void_benchmark}, and \secref{sec:void_ablation} provided quantitative results in \tabref{tab:results_void_benchmark} and \ref{tab:results_void_ablation} and qualitative results in \figref{fig:results_void_benchmark}. Here, we provide additional qualitative results in \figref{fig:void_additional_results} to show how our approach performs on a variety of scenes -- both indoor and outdoor -- from the VOID dataset. The figure is arranged in two panels of $3\times2$ grids, where each panel contains a sample RGB image (left) that is fed to our depth completion network as input, and the corresponding colored pointcloud  (right) produced by our approach, viewed at a different vantage point. The pointclouds are obtained by back-projecting the color pixels to the estimated depth. We used an input sparse depth density level of $\approx 0.5\%$ to produce the results. Our approach can provide detailed reconstructions of scenes from both indoor (e.g. right panel, last row: equipment from mechanical lab) and outdoor settings (e.g. left panel: flowers and leaves of plants in garden). It is also able to recover small objects such as the mouse on the desk in the mechanical lab, and structures at very close range (e.g. left panel, last row: staircase located less than half a meter from the camera). 

\clearpage

\begin{table}
\caption{Quantitative Pose Ablation Study KITTI Odometry Sequence 09 and 10.}
\vspace{-1em}
\begin{adjustwidth}{-.0in}{-.0in} 
\small
\centering
\begin{tabular}{l c c c c}
    \toprule
    Pose & ATE (\textit{m}) & ATE-5F (\textit{m}) & RPE (\textit{m}) & RRE (\textit{$^\circ$})\\ \midrule
    {\em Sequence 09} & & & & \\
    \midrule
    Euler     
    & 34.38 & 0.091 & 0.107 & 0.176 \\ 
    Exp.
    & 27.57 & 0.091 & 0.108 & 0.170 \\ 
    Exp. w/ Consistency
    & {\bf 18.18} & {\bf 0.080} & {\bf 0.094} & {\bf 0.157} \\ \midrule
    {\em Sequence 10} & & & & \\
    \midrule
    Euler     
    & 32.37 & 0.067 & 0.094 & 0.251 \\ 
    Exp.
    & 25.18 & {\bf 0.059} & 0.091 & 0.225 \\ 
    Exp. w/ Consistency
    & {\bf 24.60} & {\bf 0.059} & {\bf 0.081} &  {\bf 0.218} \\
    \bottomrule
\end{tabular}
\end{adjustwidth}
\begin{tablenotes}
     We perform an ablation study on our pose representation by jointly training our depth completion network and pose network on KITTI depth completion dataset and testing only the pose network on KITTI Odometry sequence 09 and 10. We evaluate the performance of each pose network using metrics described in \secref{sec:pose_eval_metrics}. While performance of exponential parameterization and Euler angles are similar on ATE-5F, and RPE, exponential outperforms Euler angles in ATE and RRE on both sequences. Our model using exponential with pose consistency performs the best.
\end{tablenotes}
\label{tab:results_kitti_odometry_ablation}
\vspace{-1em}
\end{table}

\section{Pose Ablation Study}
\label{sec:pose_ablation_study}
\begin{figure}[t]
\centering
\begin{tabular}{cc}
\includegraphics[width=0.90\textwidth]{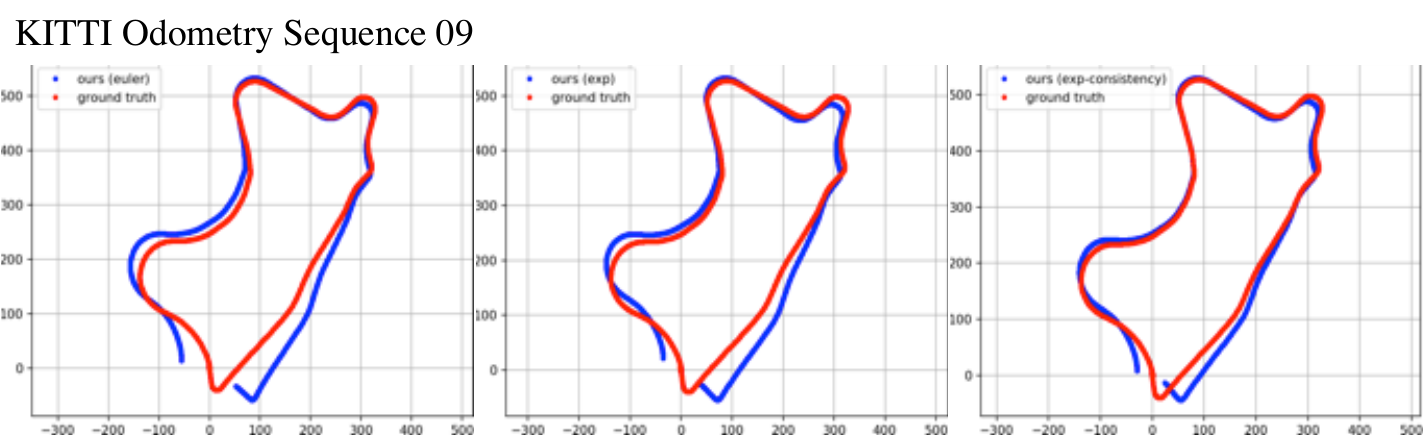} \\
\includegraphics[width=0.90\textwidth]{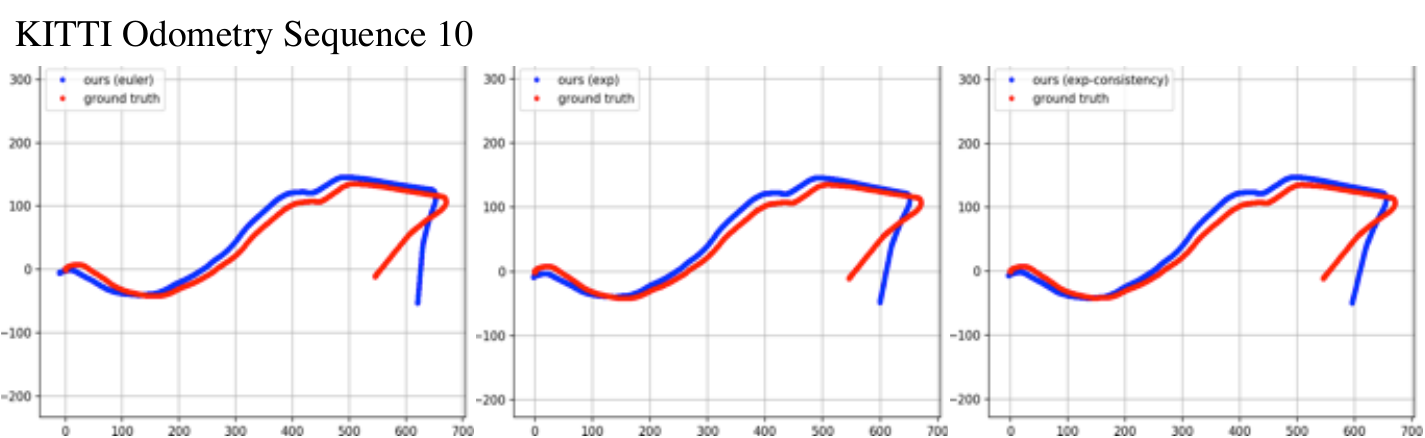}
\end{tabular}
\caption{\textit{Qualitative Pose Ablation Study KITTI Odometry Sequence 09 and 10.} We perform an ablation study on our pose representation by jointly training our depth completion network and pose network on KITTI depth completion dataset and testing only the pose network on KITTI Odometry sequence 09 and 10. We obtain the camera trajectories by chaining the pairwise camera poses estimated by our pose network. We observe that the trajectory of our method using exponential parameterization trained with pose consistency (\eqnref{eqn:pose_consistency_loss}) is most closely aligned with the ground-truth trajectory.}
\label{fig:kitti_odometry_ablation}
\end{figure}

In the main paper, we focus on the depth completion task and hence we evaluate the effects of different pose parameterizations and our pose consistency term by computing error metrics relevant to the recovery of the 3D scene on both the VOID and KITTI depth completion benchmarks. Here, we focus specifically on pose by directly evaluating the pose network on the KITTI odometry dataset in \tabref{tab:results_kitti_odometry_ablation}. We show qualitative results on the trajectory obtained by chaining pairwise camera poses estimated by each pose network in \figref{fig:kitti_odometry_ablation} and provide an analysis of the results in \secref{sec:ablation_study_kitti_odometry}.
\subsection{Pose Evaluation Metrics}
\label{sec:pose_eval_metrics}
To evaluate the performance of the pose network and its variants, we adopt two most widely used metrics in evaluating simultaneous localization and mapping (SLAM) systems: absolute trajectory error (ATE) and relative pose error (RPE) \cite{sturm2012benchmark} along with two novel metrics tailored to the evaluation of pose networks.

Given a list of estimated camera poses $\hat g^T\doteq\{\hat g_1, \hat g_2, \cdots, \hat g_T\}$, where $\hat g_t \in SE(3)$, relative to a fixed world frame, and the list of corresponding ground truth poses $g^T \doteq \{g_1, g_2, \cdots, g_T\}$, where $g_t \in SE(3)$, ATE reads
\begin{equation}
\mathrm{ATE}(\hat g^T, g^T) = \sqrt{\frac{1}{T} \sum_{t=1}^{T} \| \mathrm{trans}(g_t^{-1} \hat g_t)\|_2^2}
\end{equation}
where the function $\mathrm{trans}: SE(3)\mapsto \real^3$ extracts the translational part of a rigid body transformation. ATE is essentially the root mean square error (RMSE) of the translational part of the estimated pose over all time indices. \cite{zhou2017unsupervised} proposed a ``5-frame'' version of ATE (ATE-5F) -- the root mean square of ATE of a 5-frame sliding window over all time indices, which we also incorporate.

While ATE measures the overall estimation accuracy of the whole trajectory -- suitable for evaluating full-fledged SLAM systems where a loop closure module presents, it does not faithfully reflect the accuracy of our pose network since 1) our pose network is designed to estimate pairwise poses, and 2) thus by simply chaining the pose estimates overtime, the pose errors at earlier time instants are more pronounced. Therefore, we also adopt RPE to measure the estimation accuracy locally:
\begin{equation}
    \mathrm{RPE}(\hat g^T, g^T; \Delta) 
    = \sqrt{
    \frac{1}{T-\Delta} \sum_{t=1}^{T-\Delta}
    \| 
        \mathrm{trans} \big( (g_t^{-1}  g_{t+\Delta})^{-1} (\hat g_t^{-1} \hat g_{t+\Delta})\big) 
    \|_2^2
    }
\end{equation}
which is essentially the end-point relative pose error of a sliding window averaged over time. By measuring the end-point relative pose $\hat g_{t\tau}\doteq \hat g_t^{-1} \hat g_{t+\Delta}$, where $\tau\doteq t + \Delta$, over a sliding window $[t, t+\Delta]$, we are able to focus more on the relative pose estimator (the pose network) itself rather than the overall localization accuracy. In our evaluation, we choose a sliding window of size 1, i.e., $\Delta=1$. However, RPE is affected only by the accuracy of the translational part of the estimated pose, as we expand the relative pose error:
\begin{align}
g_{t\tau}^{-1} \hat g_{t\tau} &= (R_{t\tau}, T_{t\tau})^{-1} \cdot (\hat R_{t\tau}, \hat T_{t\tau}) \\
&= (R_{t\tau}^\top \hat R_{t\tau}, -R_{t\tau}^\top \hat T_{t\tau} + T_{t\tau})
\end{align}
leading to $\mathrm{trans}(g_{t\tau}^{-1} \hat g_{t\tau})=-R_{t\tau} \hat T_{t\tau} + T_{t\tau}$, where the rotational part $\hat R_{t\tau}$ of the estimated pose disappears! Therefore, to better evaluate the rotation estimation, and, more importantly, to study the effect of different rotation parameterization and the pose consistency term, we propose the relative rotation error (RRE) metric:
\begin{equation}
    \mathrm{RRE}(\hat g^T, g^T; \Delta) 
    = \sqrt{
    \frac{1}{T-\Delta} \sum_{t=1}^{T-\Delta}
    \| 
        \log\big( 
        \mathrm{rot} ( g_{t\tau}^{-1} \hat g_{t\tau} ) 
        \big)
    \|_2^2
    }
\end{equation}
where $\mathrm{rot}: SE(3)\mapsto SO(3)$ extracts the rotational part of a rigid body transformation, and $\log: SO(3)\mapsto \real^3$ is the logarithmic map for rotations.

\subsection{Ablation Study on KITTI Odometry}
\label{sec:ablation_study_kitti_odometry}
We perform an ablation study on the effects of our pose parameterizations and our pose consistency in \tabref{tab:results_kitti_odometry_ablation} and provide qualitative results showing the trajectory predicted by our pose network in \figref{fig:kitti_odometry_ablation}. We jointly trained our depth completion network and our pose network on the KITTI depth completion dataset and evaluate the pose network on sequence 09 and 10 of the KITTI Odometry dataset. 

For sequence 09, our pose network using exponential parameterization performs comparably to Euler angles on the ATE-5F and RPE metrics while outperforming Euler by $\approx 20\%$ on ATE and $\approx 3.4\%$ on RRE. This result suggests that while within a small window Euler and exponential perform comparably on translation, exponential is a better pose parameterization and globally more correct. We additionally see that exponential outperforms Euler angles on all metrics in sequence 10. 

Our best results are achieved using exponential parameterization with our pose consistency term (\eqnref{eqn:pose_consistency_loss}): on sequence 09, it outperformed Euler and exponential without pose consistency by $\approx 47.1\%$ and $\approx 28.9\%$ on ATE, $\approx 12.1\%$ and $\approx 13\%$ on RPE, $\approx 10.8\%$ and $\approx 7.6\%$ on RRE, respectively, and both by $\approx 12.1\%$ on ATE-5F. On sequence 10, it outperformed Euler and exponential by $\approx 24\%$ and $\approx 2.3\%$ on ATE, $\approx 13.8\%$ and $\approx 11\%$ on RPE, and $\approx 13.1\%$ and $\approx 3.1\%$ on RRE, respectively. It also beat Euler by $\approx 12\%$ on RPE and is comparable to exponential on the metric.

\clearpage

\begin{figure}[h]
    \centering
    \includegraphics[width=0.99\textwidth]{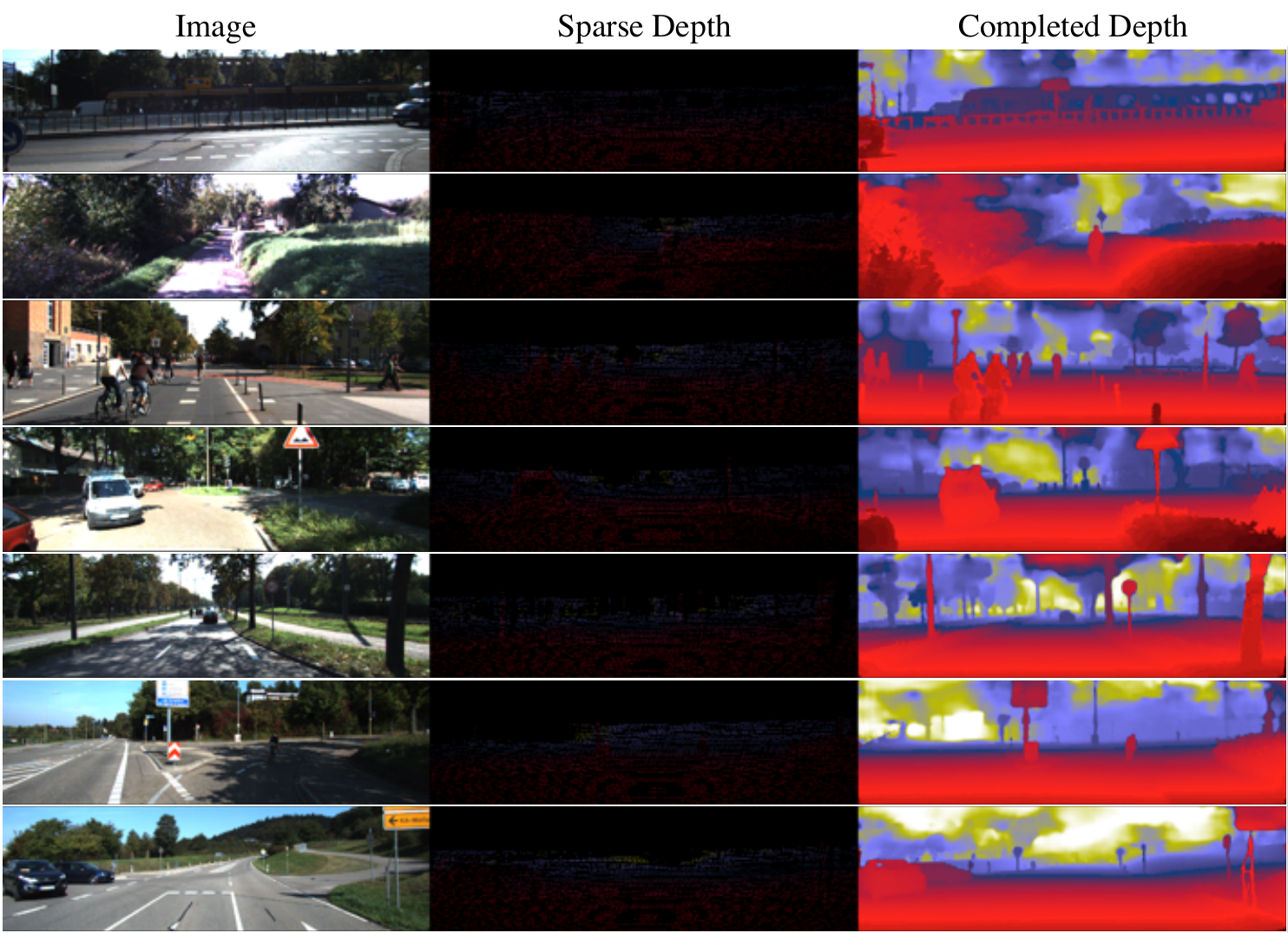}
    \caption{\textit{Qualitative Results on KITTI Depth Completion Test Set.} We show results from various scenes on the KITTI test set. The sparse depth input on the KITTI benchmark is concentrated on the lower half of the image domain. Our network learns to predict structures that do not have any sparse points (e.g. street sign in row 3, 5, and 6). Also, we are able to recover predestrians (e.g. rows 2, and 3) and thin structures well (e.g. guard rails in row 1, poles in row 3, 4, 5, and 6, and 7).}
    \label{fig:results_kitti_benchmark_additional}
\end{figure}

\section{More Results on KITTI Depth Completion Benchmark}

In the main paper, we evaluated our approach on the KITTI depth completion benchmark test set in \secref{sec:kitti_depth_completion_benchmark} and performed an ablation study on the validation set in \secref{sec:kitti_depth_completion_ablation_study}. Quantitative results are shown in \tabref{tab:results_kitti_benchmark}, \ref{tab:results_kitti_ablation} and qualitative results in \figref{fig:results_kitti_benchmark}. However, as the KITTI online depth completion benchmark only shows the first 20 samples from the test set, we provide additional qualitative results on a variety of scenes in \figref{fig:results_kitti_benchmark_additional} to better represent our performance on the test set.

The results in \figref{fig:results_kitti_benchmark_additional} were produced by our VGG11 model trained using the full loss function (\eqnref{eqn:loss_function}) with exponential parameterization for rotation. Our method is able to recover pedestrians and thin structures well (e.g. the guard rails, and street poles). Additionally, our network is also able to recover structures that do not have any associated sparse lidar points (e.g structures located on the upper half of the image domain). This can be attributed to our photometric data-fidelity term (\secref{sec:photometric_consistency_loss}). As show in \figref{fig:train_progression}, our network first learns to copy the input scaffolding and to output it as the prediction. It later learns to fuse information from the input image to produce a prediction that includes elements from the scene that is missing from the scaffolding. 

\clearpage

\begin{figure*}[h!]
    \centering
    \includegraphics[width=0.7\textwidth]{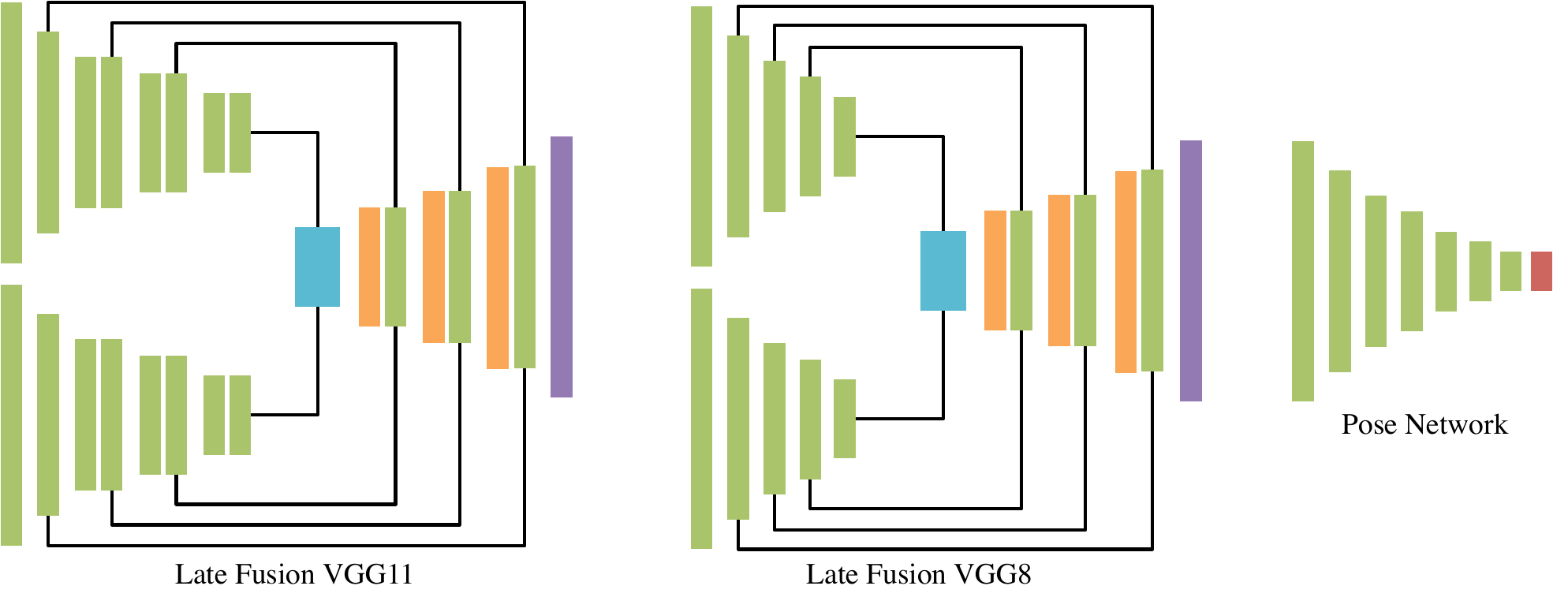}
    \caption{Network architectures. Green denotes convolution, orange deconvolution, and purple upsampling. Blue denotes the latent representation, and red the output of pose network. Our VGG11 and VGG8 architectures following the late fusion paradigm \cite{ma2018self, yang2019dense}, and our auxiliary pose network to predict relative pose between two frames for constructing our photometric and pose consistency loss (\eqnref{eqn:photometric_consistency_loss}, \ref{eqn:pose_consistency_loss}). Our auxiliary pose network is used only in training and not inference.}
    \vspace{-1.5em}
    \label{fig:networks_all}
    \vspace{-0.1em}
\end{figure*}

\section{Network Architecture}
We trained our model using two network architectures (\figref{fig:networks_all}) following the late fusion paradigm: (i) our main model using a VGG11 \cite{simonyan2014very} encoder (\tabref{tab:vgg11_encoder_architecture}), and (ii) our light weight model using a VGG8 \cite{simonyan2014very} encoder (\tabref{tab:vgg8_encoder_architecture}). Both encoders use the same decoder (\tabref{tab:decoder_architecture}). 

\begin{table*}[h!]
\caption{VGG11 Encoder Architecture}
\centering
\small
\setlength\tabcolsep{10pt}
\begin{tabular}{lcccccccc@{}}
{\bf VGG11 Encoder} & \multicolumn{2}{c}{kernel} & \multicolumn{2}{c}{channels} & \multicolumn{2}{c}{resolution} & \\ 
\cmidrule(lr){2-3} 
\cmidrule(lr){4-5}
\cmidrule(lr){6-7}
layer & size & stride & in & out & in & out & \# params & input \\ 
\midrule
{\em Image Branch} & \multicolumn{1}{l}{} & \multicolumn{1}{l}{} & \multicolumn{1}{l}{} & \multicolumn{1}{l}{} & \multicolumn{1}{l}{} & \multicolumn{1}{l}{} & \multicolumn{1}{l}{} \\
\midrule
conv1\_image  & 5 & 2 & 3   & 48  & 1    & 1/2  & $\approx$ 3.6K & image  \\ \midrule
conv2\_image  & 3 & 2 & 48  & 96  & 1/2  & 1/4  & $\approx$ 41K  & conv1\_image  \\ \midrule
conv3\_image  & 3 & 1 & 96  & 192 & 1/4  & 1/4  & $\approx$ 166K & conv2\_image  \\ \midrule
conv3b\_image & 3 & 1 & 192 & 192 & 1/4  & 1/4  & $\approx$ 331K & conv3\_image  \\ \midrule
conv4\_image  & 3 & 1 & 192 & 384 & 1/8  & 1/8  & $\approx$ 663K & conv3b\_image \\ \midrule
conv4b\_image & 3 & 1 & 384 & 384 & 1/8  & 1/8  & $\approx$ 1.3M & conv4\_image  \\ \midrule
conv5\_image  & 3 & 1 & 384 & 384 & 1/16 & 1/16 & $\approx$ 1.3M & conv4b\_image \\ \midrule
conv5b\_image & 3 & 2 & 384 & 384 & 1/16 & 1/32 & $\approx$ 1.3M & conv5\_image  \\ \midrule
{\em Depth Branch} & \multicolumn{1}{l}{} & \multicolumn{1}{l}{} & \multicolumn{1}{l}{} & \multicolumn{1}{l}{} & \multicolumn{1}{l}{} & \multicolumn{1}{l}{} & \multicolumn{1}{l}{} \\ \midrule
conv1\_depth  & 5 & 2 & 2   & 16  & 1    & 1/2  & $\approx$ 0.8K & depth  \\ \midrule
conv2\_depth  & 3 & 2 & 16  & 32  & 1/2  & 1/4  & $\approx$ 4.6K & conv1\_depth  \\ \midrule
conv3\_depth  & 3 & 1 & 32  & 64  & 1/4  & 1/4  & $\approx$ 18K  & conv2\_depth  \\ \midrule
conv3b\_depth & 3 & 1 & 64  & 64  & 1/4  & 1/4  & $\approx$ 37K  & conv3\_depth  \\ \midrule
conv4\_depth  & 3 & 1 & 64  & 128 & 1/8  & 1/8  & $\approx$ 74K  & conv3b\_depth \\ \midrule
conv4b\_depth & 3 & 1 & 128 & 128 & 1/8  & 1/8  & $\approx$ 147K & conv4\_depth  \\ \midrule
conv5\_depth  & 3 & 1 & 128 & 128 & 1/16 & 1/16 & $\approx$ 147K & conv4b\_depth \\ \midrule
conv5b\_depth & 3 & 2 & 128 & 128 & 1/16 & 1/32 & $\approx$ 147K & conv5\_depth  \\ \midrule
{\em Latent Encoding} & \multicolumn{1}{l}{} & \multicolumn{1}{l}{} & \multicolumn{1}{l}{} & \multicolumn{1}{l}{} & \multicolumn{1}{l}{} & \multicolumn{1}{l}{} & \multicolumn{1}{l}{} \\ \midrule
latent        & - & - & 384+128 & 512   & 1/32 & 1/32 & 0    & conv5b\_image $\|$ conv5b\_depth \\
\midrule 
\midrule
Total Parameters & $\approx$ 5.7M
\end{tabular}
\begin{tablenotes}Our VGG11 \cite{simonyan2014very} encoder following the late fusion paradigm \cite{jaritz2018sparse, yang2019dense} contains $\approx$ 5.7M parameters as opposed to the $\approx$23.8M and $\approx$14.8M parameters used by \cite{ma2018self} and \cite{yang2019dense}, respectively. The $\|$ symbol denotes concatenation. Resolution ratio with respect to image size.
\end{tablenotes}
\label{tab:vgg11_encoder_architecture}
\end{table*}

\begin{table*}[h!]
\caption{VGG8 Encoder Architecture}
\centering
\small
\setlength\tabcolsep{10pt}
\begin{tabular}{lcccccccc@{}}
{\bf VGG8 Encoder} & \multicolumn{2}{c}{kernel} & \multicolumn{2}{c}{channels} & \multicolumn{2}{c}{resolution} & \\ 
\cmidrule(lr){2-3} 
\cmidrule(lr){4-5}
\cmidrule(lr){6-7}
layer & size & stride & in & out & in & out & \# params & input \\ 
\midrule
{\em Image Branch} & \multicolumn{1}{l}{} & \multicolumn{1}{l}{} & \multicolumn{1}{l}{} & \multicolumn{1}{l}{} & \multicolumn{1}{l}{} & \multicolumn{1}{l}{} & \multicolumn{1}{l}{} \\
\midrule
conv1\_image  & 5 & 2 & 3   & 48  & 1    & 1/2  & $\approx$ 3.6K & image  \\ \midrule
conv2\_image  & 3 & 2 & 48  & 96  & 1/2  & 1/4  & $\approx$ 41K  & conv1\_image  \\ \midrule
conv3b\_image & 3 & 2 & 96  & 192 & 1/4  & 1/8  & $\approx$ 166K & conv2\_image  \\ \midrule
conv4b\_image & 3 & 2 & 192 & 384 & 1/8  & 1/16 & $\approx$ 663K & conv3b\_image \\ \midrule
conv5b\_image & 3 & 2 & 384 & 384 & 1/16 & 1/32 & $\approx$ 1.3M & conv4b\_image \\ \midrule
{\em Depth Branch} & \multicolumn{1}{l}{} & \multicolumn{1}{l}{} & \multicolumn{1}{l}{} & \multicolumn{1}{l}{} & \multicolumn{1}{l}{} & \multicolumn{1}{l}{} & \multicolumn{1}{l}{} \\ \midrule
conv1\_depth  & 5 & 2 & 2   & 16  & 1    & 1/2  & $\approx$ 0.8K & depth  \\ \midrule
conv2\_depth  & 3 & 2 & 16  & 32  & 1/2  & 1/4  & $\approx$ 4.6K & conv1\_depth  \\ \midrule
conv3b\_depth & 3 & 1 & 32  & 64  & 1/4  & 1/4  & $\approx$ 18K  & conv2\_depth  \\ \midrule
conv4b\_depth & 3 & 1 & 64  & 128 & 1/8  & 1/16 & $\approx$ 74K  & conv3b\_depth \\ \midrule
conv5b\_depth & 3 & 2 & 128 & 128 & 1/16 & 1/32 & $\approx$ 147K & conv4b\_depth \\ \midrule
{\em Latent Encoding} & \multicolumn{1}{l}{} & \multicolumn{1}{l}{} & \multicolumn{1}{l}{} & \multicolumn{1}{l}{} & \multicolumn{1}{l}{} & \multicolumn{1}{l}{} & \multicolumn{1}{l}{} \\ \midrule
latent        & - & - & 384+128 & 512   & 1/32 & 1/32 & 0    & conv5b\_image $\|$ conv5b\_depth \\
\midrule 
\midrule
Total Parameters & $\approx$ 2.4M
\end{tabular}
\begin{tablenotes}
    Our light-weight VGG8 \cite{simonyan2014very} encoder following the late fusion paradigm \cite{jaritz2018sparse, yang2019dense} contains only $\approx$ 2.4M parameters as opposed to the $\approx$23.8M and $\approx$14.8M parameters used by \cite{ma2018self} and \cite{yang2019dense}, respectively. The $\|$ symbol denotes concatenation. Resolution ratio with respect to image size. Note that our light-weight model performs similarly to our VGG11 model.
\end{tablenotes}
\label{tab:vgg8_encoder_architecture}
\end{table*}

\begin{table*}[h!]
\caption{Decoder Architecture}
\centering
\small
\setlength\tabcolsep{8pt}
\begin{tabular}{lcccccccc@{}}
{\bf Decoder} & \multicolumn{2}{c}{kernel} & \multicolumn{2}{c}{channels} & \multicolumn{2}{c}{resolution} & \\ 
\cmidrule(lr){2-3} 
\cmidrule(lr){4-5}
\cmidrule(lr){6-7}
layer & size & stride & in & out & in & out & \# params & input \\ 
\midrule
deconv5 & 3 & 2 & 512         & 256 & 1/32 & 1/16  & $\approx$ 1.2M & latent   \\ \midrule
concat5 & - & - & 256+384+128 & 768 & 1/16 & 1/16  & 0              & deconv5$\|$conv4b\_image$\|$conv4b\_depth \\ \midrule
conv5   & 3 & 1 & 768         & 256 & 1/16 & 1/16  & $\approx$ 1.8M & concat5  \\ \midrule
deconv4 & 3 & 2 & 256         & 128 & 1/16 & 1/8   & $\approx$ 295K & conv5    \\ \midrule
concat4 & - & - & 128+192+64  & 384 & 1/8  & 1/8   & 0           & deconv4$\|$conv3b\_image$\|$conv3b\_depth \\ \midrule
conv4   & 3 & 1 & 384         & 128 & 1/8  & 1/8   & $\approx$ 442M & concat4  \\ \midrule
deconv3 & 3 & 2 & 128         & 128 & 1/8  & 1/4   & $\approx$ 147K  & conv4    \\ \midrule
concat3 & - & - & 128+96+32   & 256 & 1/4  & 1/4   & 0              & deconv3$\|$conv2\_image$\|$conv2\_depth \\ \midrule
conv3   & 3 & 1 & 256         & 64  & 1/4  & 1/4   & $\approx$ 147K & concat3  \\ \midrule
deconv2 & 3 & 2 & 64          & 64  & 1/4  & 1/2   & $\approx$ 37K  & conv3    \\ \midrule
concat2 & - & - & 64+48+16    & 128 & 1/2  & 1/2   & 0              & deconv2$\|$conv1\_image$\|$conv1\_depth \\ \midrule
conv2   & 3 & 1 & 128         & 1   & 1/2  & 1/2   & $\approx$ 1.2K & concat2  \\ \midrule
output  & - & - & -           & -   & 1/2  & 1     & 0              & $\uparrow$ conv2  \\
\midrule 
\midrule
Total Parameters & $\approx$ 4M
\end{tabular}
\begin{tablenotes}
    \raggedright
    Our decoder contains $\approx$ 4M parameters. The $\|$ symbol denotes concatenation and the $\uparrow$ symbol denotes upsampling. Resolution ratio with respect to image size.
\end{tablenotes}
\label{tab:decoder_architecture}
\end{table*}

\clearpage

\noindent{\bf Depth completion networks.} Our VGG11 and VGG8 model (\figref{fig:networks_all}) contain a total of $\approx 9.7$M and $\approx 6.4$M parameters, respectively. In comparison to \cite{ma2018self} with $\approx 27.8$M parameters and \cite{yang2019dense} with $\approx 18.8$M, our VGG11 model have a $65.1\%$ and $48.4\%$ reduction in parameters over \cite{ma2018self} and \cite{yang2019dense}, respectively; our VGG8 model have a $80\%$ and $66\%$ reduction over \cite{ma2018self} and \cite{yang2019dense}. The image and depth branches of the encoder process the image and depth inputs separately -- weights are not shared. The results of the encoders are concatenated as the latent representation and passed to the decoder for depth completion. The decoder makes the prediction at 1/2 resolution. The final layer of the decoder is an upsampling layer. 

\noindent{\bf Pose Network.} Our pose network takes a pair of images as input and regresses the relative pose between the images. Reversing the order of the image will reverse the relative pose as well. We take the average across the width and height dimensions of the pose network output to produce a 6 element vector. We use 3 elements to model rotation and the rest to model translation. 

\noindent{\bf Including Pose Network in Total Parameters.} We follow the network parameter computations of \cite{yang2019dense} who employs an additional network trained on ground truth for regularization during training. Our pose network (\tabref{tab:posenet_architecture}) is an auxiliary network that is only used in training, and not during inference.  Hence, we do not include it in the total number of parameters. However, even if we do, our pose network has $\approx$ 1M parameters, making our total for VGG11 to be $\approx$ 10.7M and VGG8 to be $\approx$ 7.4M. Our VGG11 model is still has a $61.5\%$ reduction in parameter, and our VGG8 a $73.4\%$ over the 27.8M parameters used by \cite{ma2018self}. If we include the auxiliary prior network of \cite{yang2019dense}, containing 10.1M parameters, that is used for regularization during training, then \cite{yang2019dense} has a total of 28.8M parameters. Our VGG11 model, therefore, has a $62.8\%$ reduction in parameters over \cite{yang2019dense} and our VGG8 has a $74.3\%$ reduction.

\begin{table*}[t!]
\caption{Pose Network Architecture}
\centering
\small
\setlength\tabcolsep{15pt}
\begin{tabular}{lcccccccc@{}}
{\bf Pose Network} & \multicolumn{2}{c}{kernel} & \multicolumn{2}{c}{channels} & \multicolumn{2}{c}{resolution} & \\ 
\cmidrule(lr){2-3} 
\cmidrule(lr){4-5}
\cmidrule(lr){6-7}
layer & size & stride & in & out & in & out & \# params & input \\ 
\midrule
conv1  & 7 & 2 & 6   & 16  & 1     & 1/2   & $\approx$ 4.7K & image pair \\ \midrule
conv2  & 5 & 2 & 16  & 32  & 1/2   & 1/4   & $\approx$ 13K  & conv1      \\ \midrule
conv3  & 3 & 2 & 32  & 64  & 1/4   & 1/8   & $\approx$ 18K  & conv2      \\ \midrule
conv4  & 3 & 2 & 64  & 128 & 1/8   & 1/16  & $\approx$ 74K  & conv3      \\ \midrule
conv5  & 3 & 2 & 128 & 256 & 1/16  & 1/32  & $\approx$ 295K & conv4      \\ \midrule
conv6  & 3 & 2 & 256 & 256 & 1/32  & 1/64  & $\approx$ 295K & conv5      \\ \midrule
conv7  & 3 & 2 & 256 & 256 & 1/64  & 1/128 & $\approx$ 295K & conv6      \\ \midrule
output & 3 & 1 & 256 & 6   & 1/128 & 1/128 & $\approx$ 14K  & conv7      \\
\midrule 
\midrule
Total Parameters & $\approx$ 1M
\end{tabular}
\begin{tablenotes}
\raggedright
Our auxiliary pose network contains $\approx$ 1M parameters and is only used during training to construct the photometric and pose consistency loss (\eqnref{eqn:photometric_consistency_loss}, \ref{eqn:pose_consistency_loss}). The output is averaged along its width and height dimensions to result in a 6 element vector -- of which 3 elements are used to compose rotation and the rest for translation.
\end{tablenotes}
\label{tab:posenet_architecture}
\end{table*}

\end{appendices}

\end{document}